\documentclass[final]{IEEEtran}

\usepackage{cite}
\usepackage{url}
\usepackage{amsmath,amssymb,amsfonts}
\usepackage{algorithmic}
\usepackage{graphicx}
\usepackage{textcomp}
\usepackage{xcolor}
\usepackage{booktabs} 
\def\BibTeX{{\rm B\kern-.05em{\sc i\kern-.025em b}\kern-.08em
    T\kern-.1667em\lower.7ex\hbox{E}\kern-.125emX}}

\newcommand{\cready}[1]{{}}

\newcommand{\danny}[1]{\textbf{\color{blue}Danny: #1}}

\newcommand{\remove}[1]{}
    
\begin{document}

\title{Lossless and Near-Lossless Compression for Foundation Models}


\author{Moshik Hershcovitch$^{1,2}$, Leshem Choshen$^{1,3}$, Andrew Wood$^{4}$, \\ Ilias Enmouri$^{1}$, Peter Chin$^{5}$, Swaminathan Sundararaman$^{1}$, Danny Harnik$^{1}$\\
$^1$ IBM Research 
$^2$ Tel-Aviv University
$^3$ MIT \\
$^4$ Boston University 
$^5$ Dartmouth University \\
}

\maketitle

\begin{abstract}
With the growth of model sizes and scale of their deployment, their sheer size burdens the infrastructure requiring more network and more storage to accommodate these. While there is a vast literature about reducing model sizes, we investigate a more traditional type of compression -- one that compresses the model to a smaller form and is coupled with a decompression algorithm that returns it to its original size -- namely lossless compression. 

Somewhat surprisingly, we show that specific lossless compression can gain significant network and storage reduction on popular models, at times reducing over $50\%$ of the model size.  We investigate the source of model compressibility and introduce specialized compression variants tailored for models that further increase the effectiveness of compression. 

We also categorize models to compressibility groups and introduce a tunable lossy compression technique that can further reduce size even on the group of less compressible models with little to no effect on the model accuracy. Finally, we explore the usefulness of delta compression for checkpointing and model variations.  

We estimate that these methods could save over an ExaByte per month of network traffic downloaded from a large model hub like Hugging~Face.  


\end{abstract}

\section{Introduction}
\label{submission}

With scale, we have learned that models gain performance and with it, gain popularity.
With scale, models also require more memory and with popularity more communication bandwidth. Taken together, we observe strains on communication bottlenecks that call for efficient solutions. Storage requirements, while often ignored, may accumulate to hundreds or thousands of times the size of a model if checkpoints~\cite{Biderman2023PythiaAS} or distributed updates are to be saved (c.f.,~\ref{sec:related}) \cite{kandpal2023git,don-yehiya-etal-2023-cold,zhang2021survey}. 

Similarly, models are repeatedly moved around in multiple channels: from a storage hub to inference machines; from training/fine-tuning nodes to the storage backend; between GPU nodes during distributed training and so on.
Network hubs epitomize the strains by model size. For instance, with over  14.5 GBs and 2.77 M downloads per month from Hugging~Face \cite{Wolf2019HuggingFacesTS} Mistral \cite{jiang2023mistral} alone requires 40 PBs of transferred information a month.

A large body of work has been aimed at reducing model sizes focusing on the number of computations in inference. Such methods transform the model into a smaller one in an irreversible fashion. For example, distillation \cite{gou2021knowledge}, pruning  \cite{ma2021effective} and quantization \cite{gholami2021survey_quantization} either remove nodes from the network or reduce each parameter size. Since these methods main focus is on inference speed, they are bound to create a format of an actual running model. As such, they don't necessarily push the space saving to its limit, and are not stored in the minimal possible way.  

\begin{table*}[t]
\caption{Top ranked downloaded models from Hugging~Face and the potential traffic savings if compressed.}
\label{table:HF}
\vskip 0.15in
\begin{center}
\begin{small}
\begin{sc}
\begin{tabular}{lccccc}
\toprule
Model &  Model &  \#Monthly & Rank & Compression & Potential \\
name &  Size & Downloads & & Ratio & Savings\\
\hline \hline
\textbf{Wav2vec} & 1.26 GB & 63M &\#1 & \textbf{85.2\%} & 11.7 PB 
\\
\textbf{Bert} & 0.4 
 GB & 43.7M& \#2 & \textbf{85.3\%} & 2.6 PB
\\ 
\textbf{RoBERTa} &  0.5 GB & 15M & \#3 & \textbf{47.0\%} & 4 PB  
\\
\textbf{GPT2} &  0.5 
 GB & 14.6M & \#4 & \textbf{78.1\%} & 1.6 PB
\\
\textbf{Clip} & 1.7 GB & 14.3M & \#5 & \textbf{50.1\%} &  12.2 PB 
\\
\textbf{Mistral} & 14.5 GB& 2.77M &\#\textasciitilde60 & \textbf{71.0\%} & 11.6 PB 
\\
\textbf{Bloom} & 328.2 GB& 278K &\#100+ & \textbf{71.4\%} & 26.1 PB 
\\
\bottomrule
\end{tabular}
\end{sc}
\end{small}
\end{center}
\end{table*}

In this work, on the other hand, we follow a more traditional definition of compression typically used for networking and storage. Compression that is also accompanied by a decompression process, returning a model to its original size and usability. 
This definition encompasses among other things all forms of lossless compression.  

Surprisingly, we observe (\S\ref{sec:compression}) that even standard lossless compressors like zlib \cite{deutsch1996zlib} or zstd \cite{collet2018zstandard} can achieve meaningful savings and these can be further amplified using specialized modifications to the compressors. While common rationale expects model parameters to have high entropy and therefore be non-compressible, we find that in reality there is ample redundancy in representation.  
We classify popular models from Hugging~Face \cite{Wolf2019HuggingFacesTS} into three categories with distinct compressibility traits helping to understand when and for what models it is beneficial to employ lossless compression. 

After exploring the source of compressibility in models we introduce {\em byte grouping} -- an adaptation that is tailored for the models use case (\S\ref{sec:losseless_res}). The method rearranges the bytes in a model to compress the different bytes of the parameters together. This results in grouping of similar bytes which in turn yields better compression.

We make another key observation, that fine-tuning of models often degrades their compressibility (at times significantly). This high entropy in the parameters often stems from minuscule updates. To overcome this, we introduce a novel tunable lossy compression method that can significantly improve compression ratio with no measurable harm to model accuracy. In a nutshell, this technique allows for incurring controlled inaccuracies to parameters, under the assumption that a lot of the entropy in model weights is actually redundant, i.e., noise saved to disk. 
Surprisingly, we find ranges where those precision reductions can even slightly benefit the model, corroborating a few similar findings (see \S\ref{sec:related}). While the goal of this work is not to affect models, this unexpected phenomenon is worth mentioning.

Finally, we explore the benefits of delta compression and show that by compressing the delta between two similar models one can achieve compression far greater than compressing a standalone model. This is useful for checkpointing and management of model variations.

\section{Background}
\subsection{Motivation - use cases}
With small models weighing about a Gigabyte \cite{Devlin2019BERTPO} and large ones Terrabytes \cite{Fedus2021SwitchTS} storage by itself is an issue for many purposes. However, common use cases require many model types or model versions and hence increased resources. We list some below as a motivation.

\subsubsection{Model Hubs} Large model repositories or hubs like Model Zoo \cite{modelZoo}, PyTorch \cite{Pytorch_2019}, Tensorflow \cite{tensorflowHub}, Adapter \cite{adapterhub}, Hugging~Face \cite{Wolf2019HuggingFacesTS}, IBM watsonx.data~\cite{watsonxdata} and Qualcomm® AI Hub~\cite{QualcommAIHUB} hold a large number of models and serve numerous download requests of popular models. As of 2024, Hugging~Face, the largest model hub, transfers PetaBytes of data every day, primarily downloaded data. Table~\ref{table:HF} shows some of the top ranked models\footnote{Rank as of August 2023 except for Mistral taken in March 2024.} and their compression ratio (using the methods described in Section~\ref{sec:lossless}). As seen, the potential traffic savings from compression is substantial. Note that the same trends also apply to models that are not downloaded as often, for example, the Bloom model offers significant savings due to its large initial size. 

In this use case, there are three ways in which compression can be beneficial, the first and the most important is to reduce the amount of data transferred, the second is to reduce the amount of data stored and the third is to reduce the time to download and upload those models.


\subsubsection{Distributed Training}  transfer data between nodes during training to overcome the need to save the full model and computation on a single GPU/node. In some methods, only the model weights are transferred between nodes and in other methods, the optimizer weights and gradients are transferred as well \cite{zhao2023pytorch}. Either way, distributed training is usually limited by data transfer between nodes, so compressing this data can help train larger models. 

\subsubsection{Decentralized algorithms}
Along the different ways to distribute the computation along nodes. Some methods propose to alternate training or the models themselves to accommodate different contributors training the same model. This stems from federated learning that contributes gradients \cite{zhang2021survey}, to contributing partially trained models \cite{li2022branch}, from changing the kinds of updates done \cite{lialin2023relora} to relying on volunteer computing \cite{NEURIPS2021_41a60377}, or even relying on different objectives and expertise all merged into the same model \cite{don-yehiya-etal-2023-cold}. All of those methods inherently transfer and store a lot of model versions, which also drove dedicated version control frameworks \cite{kandpal2023git},

\subsubsection{Checkpoints and Versions} During model creation, multiple intermediate versions of the models are commonly saved. This often includes tests on the training regime such as hyperparameter tuning \cite{Turner2021BayesianOI}. Even during the training of a single model, the current model is periodically checkpointed to recover after a crash, to select the best checkpoint from a few options \cite{Dodge2020FineTuningPL}, for analysis \cite{Biderman2023PythiaAS}, improve performance \cite{JunczysDowmunt2018MarianFN,Sandler2023TrainingTM} etc. 
Even though saving during checkpointing rarely slows the training time, it does burden the networking and storage, limiting the frequency and amount of saved and shared checkpoints, which are encouraged by the community (e.g.; \cite{Biderman2023PythiaAS,liu2023llm360}). 

\remove{
\begin{figure}[ht]
\vskip 0.2in
\begin{center}
\centerline{\includegraphics[width=\columnwidth]{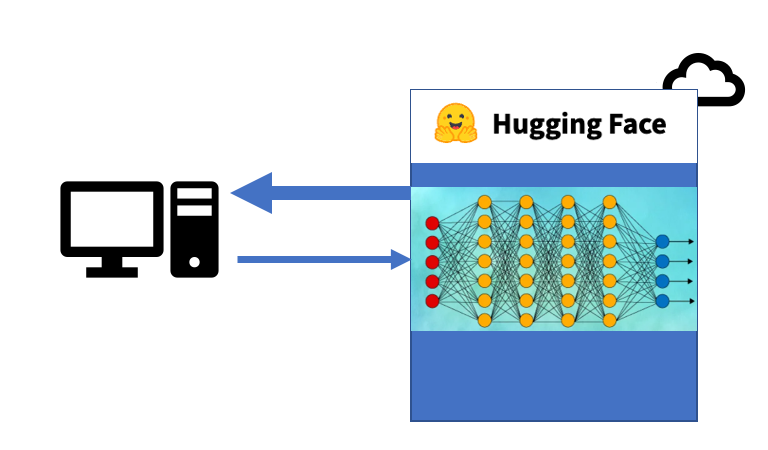}}
\caption{Model HUB}
\label{AI HUB}
\end{center}
\end{figure}
}

\subsection{Models Structure and Types}

\begin{figure}[ht]
\vskip 0.2in
\begin{center}
\centerline{\includegraphics[width=\columnwidth]{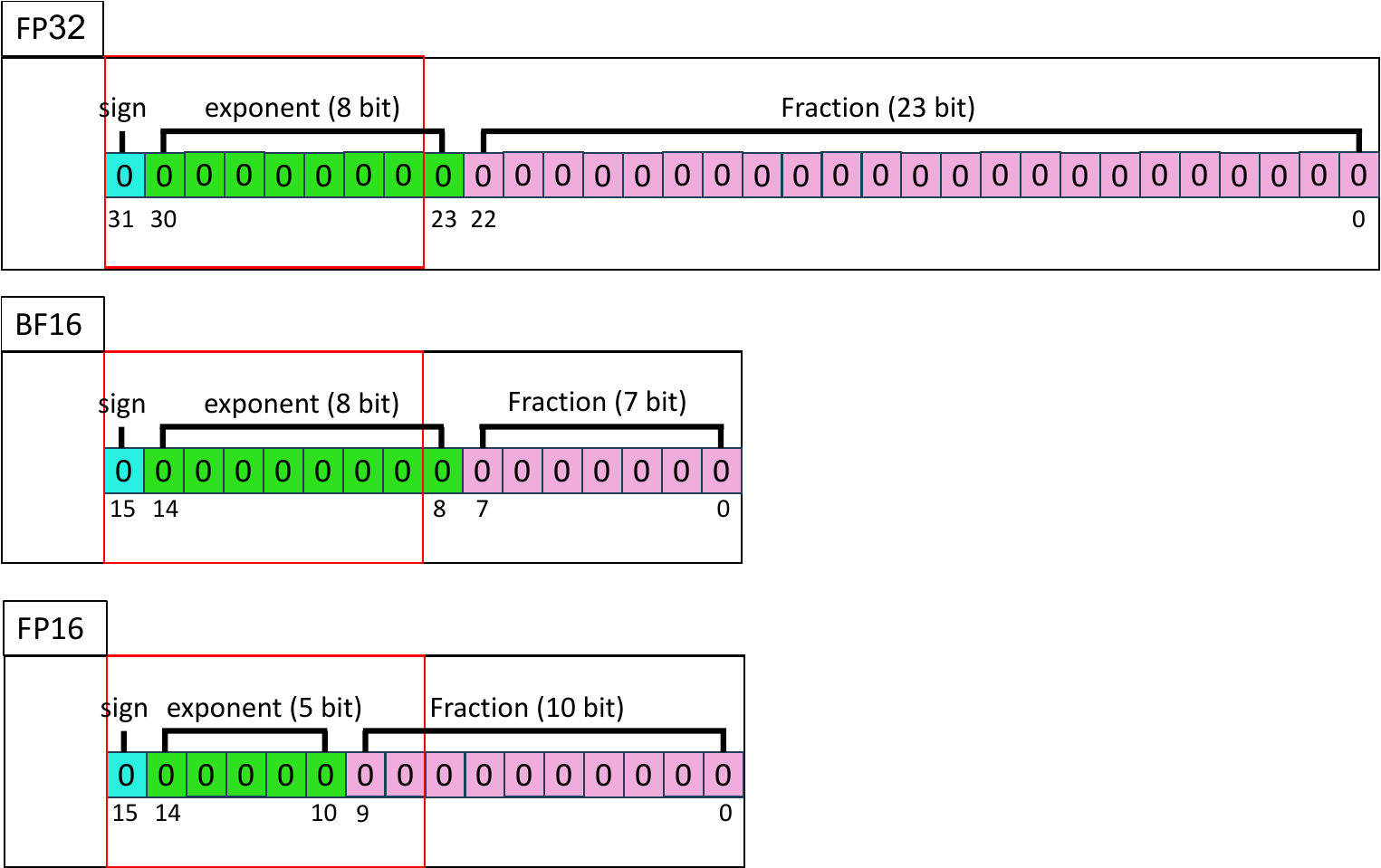}}
\caption{FP32, a sign bit + 8 bits exponents + 23 bits of mantissa. BF16, a sign bit + 8 bits exponents + 7 bits of mantissa. FP16, a sign bit + 5 bits exponents + 10 bits of mantissa}
\label{fig:FP32}
\end{center}
\end{figure}

\paragraph{Models.}\label{sec:models} Regardless of the architecture, current models are mainly a function of many matrices or tensors of different sizes and a code that can read the parameters in the matrix and convert it to a function. While a layer may contain several such tensors, for brevity we call each tensor a layer. We note that the code is negligible in weight and hence the main focus of our compression is reduced to tensors, or even put more simply, long arrays of numeric parameters.

One main variant in models that strongly affects compression tendencies is the type of parameters that make up a model. 
Parameters typically represent real numbers and as such the most straightforward (and popular) approach is to hold them using {\em floating point} numbers. Floating point is a way to represent real numbers using a fixed number of bits but a flexible scale that allows for larger numerical ranges. In a nutshell, floating point contains an exponent part - representing the range in which the real number lies, and a {\em mantissa} or {\em fraction} representing the actual number within this range. To this, a sign bit is added indicating whether a number is positive. For example, FP32 is a 32 bit floating point number with a sign bit, an 8 bit exponent and a 23 bit mantissa (see Figure~\ref{fig:FP32}). 
The real number is calculated by $(-1)^{sign} \cdot 2^{exponent-127} \cdot 1.fraction$. 
Another popular parameter type used for models is BF16~\cite{wang2019bfloat16} which simply cuts the tail end of the fraction (hence reducing the precision level) but maintains the same exponent as shown in Figure~\ref{fig:FP32}.



\section{Compression for Models}\label{sec:compression}
In this section, we introduce the basics of lossless compression variants that are relevant to model and present our variant of lossy compression.  
Note that in this paper, we evaluated only compression run in the CPU, since the GPU computations and GPU memory are more valuable resources, whereas CPU computations and CPU memory are typically abundant. 

\subsection{Lossless compression}\label{sec:lossless}
Lossless compressors are the traditional form of compression and are widely used for reducing network and storage overheads in all fields of computing. They consist of two algorithms -- compression and decompression where after applying both in sequence the output returns to the exact same state. There are countless compression techniques, those vary in the tradeoff between compressibility and compression/decompression time (see for example; \cite{squash}).

Throughout the paper, we measure the {\em compression ratio}. Namely, the percentage of the data that is left after compression -- {\em lower is better}. For example, if the method compresses a GB into a quarter of a GB it has a compression ratio of $25\%$.  

The main techniques employed in lossless compression are based on repetition removal (stemming from the seminal work of \cite{LZ77}) and entropy encoding (e.g. \cite{Huffman1952, Riss1976AC}) which reduces the entropy seen at a byte level and represents such bytes at a bit level granularity. 

We observed that compressors that employ solely repetition removal (such as LZ4; \cite{lz4}) are not very beneficial when compressing models. This is expected since to remove a repetition a span of multiple parameters should repeat itself. However, model tensors are not structured and parameters typically do not have an affinity with their neighbours, making repetitions that span multiple parameters scarce. For the rest of our experiments, we choose compressors that also use entropy encoding such as Zlib~\cite{zlib} or Zstd~\cite{zstd}. In our experiments, we chose Zstd as the underlying compressor/decompressor due to its superior speed vs.\ compression tradeoff \cite{zstd, squash}.
Note that once Byte Grouping (see \S\ref{sec:byte_grouping_background}) is used, then LZ4 fairs much better, but still Zstd is far superior in terms of compression ratio. For example, for RoBERTa we get $47.0\%$ with Zstd vs.\ $56.7\%$ with LZ4 and for Bloom we get $71.4\%$ with Zstd as opposed to $80.5\%$ with LZ4. 

\begin{table*}[t]
\caption{Compression Ratio of Models after Zstd compression with Byte Grouping.}
\label{table:loseless}
\vskip 0.15in
\begin{center}
\begin{small}
\begin{sc}
\begin{tabular}{lcccccr}
\toprule
Model & Param & Model & Compression  &  Compression Ratio\\
name & Type & Size &  Ratio & Per Byte Group  \\
\hline \hline
\textbf{Wav2vec} & FP32 & 1.2 GB &  \textbf{85.2\%} & (42.9\%, 99.0\%, 99.0\%, 98.6\%) 
\\
\textbf{Bert} & FP32 & 0.4 GB &  \textbf{85.3\%} &  (41.2\%, 99.0\%, 99.0\%, 99.0\%) 
\\ 
\textbf{GPT2} & FP32 & 0.5 GB & \textbf{78.1\%} &  (38.9\%, 90.8\%, 90.8\%, 90.8\%) 
\\
\textbf{RoBERTa(1 epoch)} & FP32 & 0.5GB &  \textbf{80.7\%} &  (42.9\%, 99.9\%, 93.5\%, 86.4\%) 
 &
\\
\textbf{RoBERTa(9 epochs)} & FP32 & 0.5GB &  \textbf{82.5\%} & (42.9\%, 99.9\%, 96.9\%, 90.2\%) 
 &
\\
 \textbf{stable-video-diffusion} & FP16 &  4.27
 GB& \textbf{84.9\%} & (69.8\%, 100\%) 
 \\
\textbf{CapybaraHermes-Mistral} & FP16 & 14.5 
 GB& \textbf{83.7\%} & (68.7\%, 98.7\%) 
)
\\
\hline
\textbf{RoBERTa} & FP32 & 0.5 GB &  \textbf{47.0\%} &  (42.9\%, 99.9\%, 44.7\%, 0.005\%) 
 &
\\
\textbf{XLM-RoBERTa} & FP32 & 1.1 GB &  \textbf{45.7\%} &  (42.6\%, 95.7\%, 44.6\%, 0.002\%) 
 &
\\

\textbf{Clip} & FP32 & 1.7 GB & \textbf{50.1\%} &  (42.1\%, 99.0\%, 49.0\%, 8.0\%) 

 \\ 
\textbf{T5 base} & FP32 & 0.8 
 GB & \textbf{35.7\%} & (42.6\%, 99.9\%, 0.005\%, 0.005\%) 
\\
\textbf{llama-13B} & FP16 &  26
 GB & \textbf{66.8\% } & ( 69\%, 64\%) 
\\ 
\textbf{tulu-7B} & FP16 & 13.5 
 GB & \textbf{66.6\% } & (68.9\%, 64\%) 
\\ 
\hline
\textbf{Falcon-7b} & BF16 & 14.4 
 GB& \textbf{71.3\%} & (42.7\%, 100\%) 
)
\\
\textbf{Bloom} & BF16 & 328.2 
 GB&  \textbf{71.4\%} & (42.3\%, 100\%) 
 \\
\textbf{openllama-3B} & BF16 & 6.9 
 GB & \textbf{71\% } & (42.1\%, 100\%) 
\\ 
\textbf{Mistral} & BF16 & 14.5 
 GB & \textbf{71\% } & (42.0\%, 100\%) 
\\ 
\hline

\bottomrule
\end{tabular}
\end{sc}
\end{small}
\end{center}
\end{table*}
\subsubsection{Understanding Model Compressibility and Byte Grouping}\label{sec:byte_grouping_background}
Initially, one may expect models to be non-compressible and show high entropy, as parameters may encode unpredictable information and differ from each other. This is correct to a certain degree, but in reality, the actual range in which parameters reside is typically limited, which reduces the entropy and opens the door for compression to be effective. 

Take a model with FP32 parameters for example. It will have high entropy in its fraction (or mantissa), but relatively low entropy in the exponent, as the parameter scales are quite limited. Therefore it makes sense to compress exponents and mantissa bytes separately.  
Indeed, compressing the exponent bytes without the interference of the mantissa bytes yields higher compressibility.

\begin{figure}[ht]
\begin{center}
\centerline{\includegraphics[width=\columnwidth]{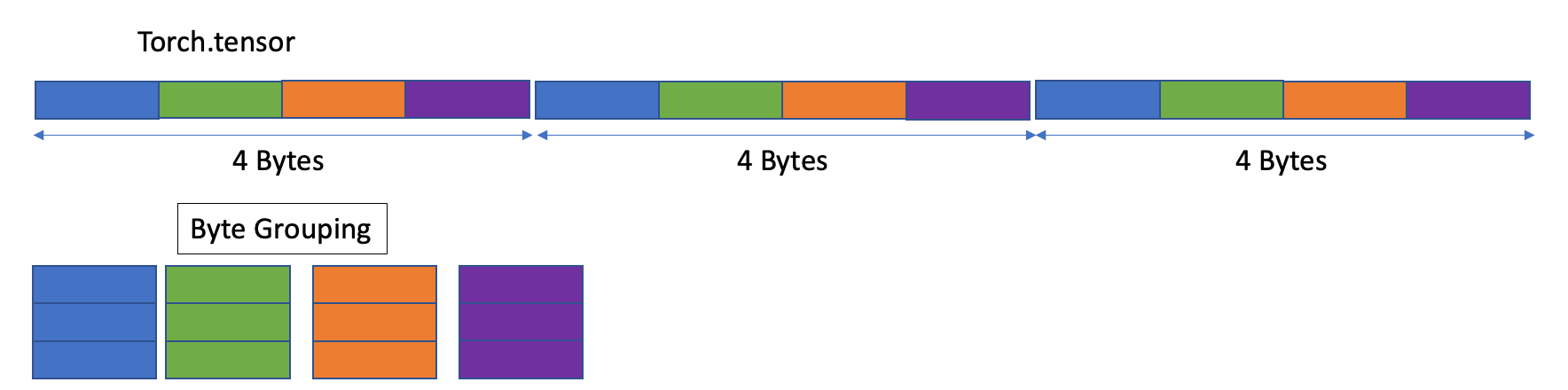}}
\caption{An example for Byte Grouping, each parameter has 4 bytes and we group them into 4 arrays.}
\label{fig:BG}
\end{center}
\end{figure}

We suggest {\bf Byte Grouping} which groups together bytes from the same position in all the model’s parameters. If each parameter in the model consists of several bytes (typically 2 or 4 bytes), then group together the first byte from all parameters, then the second byte, etc., as shown in Figure~\ref{fig:BG}. 
Note that since typical models use the same parameter type throughout the model, then byte grouping can be done without real knowledge on the model structure (except parameter type) and for example can be executed even on models that are already stored as a binary file. 

Our test shows that byte grouping the data before compressing it can improve the compression ratio between 7\%-30\% (see Section~\ref{sec:losseless_res}). 

\subsubsection{The Sign Bit} 
Another observation is that the sign bit tends to hold high entropy and that compressing it together with the exponent byte interferes with compression effectiveness. To overcome this, we consider the following approach to deal with the sign bits: translate the stream into an unsigned stream (for example, using the {\em abs} function) and store the sign values separately. The unsigned values are then fed into the compressor. This method can further improve the compression ratio without further affecting the precision. However, with lossless compression our experiments showed that the compression benefit is relatively low - on the order of $1\%$, making it unappealing due to its computational overhead.

\subsection{Model compressibility with Lossless Compression}\label{sec:losseless_res}
While model compressibility has high variance, we observe that there are essentially three popular categories of models from a compressibility standpoint.  
Table~\ref{table:loseless}  shows examples of models from the three various categories, all compressed using Zstd with its default setting (level 3) and employing the Byte Grouping technique.

The first category of models are mainly compressible in the exponent and hence have more modest savings, on the order of $15$-$20\%$ (namely with a compression ratio of $\sim$$ 80$-$85\%$). Those models are saved in FP32 of FP16. The main source of compressibility for these models is the exponent byte which is highly compressible (around $40\%$), but the other bytes hardly compress at all.   

 The second category includes ``clean" models, or base models. These have high compressibility stemming from both the exponent and the two lower bytes of the mantissa. The second byte, in all cases, is incompressible and holds most of the model's entropy. 
 Overall these models show very high compressibility (reducing the size by $50$-$65\%$) and prove very attractive for compression. Note that some of the most downloaded models from Hugging~Face fall into this category.  This leaves the two lower bytes zeroed.  We call these clean models because after fine tuning they lose much of their compressibility. For example, We fine tuned the RoBERTa model (using the Rotten Tomatos public dataset \cite{rottenTomatos}) and see that even after a single epoch of fine-tuning the model falls into the first category and only saving $20\%$. After 9 epochs of fine-tuning the compressibility is even slightly worse. 
The source of compressibility in the clean models stems from using only parts of the available entropy during the training of these models. For example, the T5 model is trained in a 16-bit environment and then cast into an FP32 parameter for further fine-tuning \cite{raffel2020exploring}. In other models the lower bytes do have some entropy in them, but not as high as the random looking bytes. 
 
 
 The final category is of BF16 models that show $\sim$$30\%$ space savings. Like the first group, the exponent is very compressible, and the mantissa is not, but in these models, the savings are more significant as the exponent makes up a larger part of the model. 

In the evaluation above we used mainly highly downloaded models from the Hugging~Face hub to check for realistic scenarios (see full model list in Appendix~\ref{app:all_models}). We include models from different modalities, sizes, architectures and saving float formats. We split them to the 3 groups mentioned above. 

To account for how commonly each model was downloaded or what version was tested (models are rarely changed after upload, but technically one can commit an update) all numbers were gathered in August 2023 except for the Mistral model and the FP16 models which were updated in March 2024.











\paragraph{The benefit of Byte Grouping}\label{sec:byte_grouping_res}
Byte grouping benefits also vary between the categories. Byte grouping reduces the size of the compressed model by $7$-$8.2\%$ on the first category, by $19$-$27\%$ on the clean model category and by $8.5$-$10\%$ on the BF16 category. 
We note that if LZ4 was used then the effect of byte grouping is massive and without byte grouping these compressors manage nearly no compression at all. For example, LZ4 on RoBERTa fails badly and achieves only $95\%$ compression ratio, but with byte grouping, this jumps to $56\%$. 






\subsection{Tunable Lossy Compression}\label{sec:lossy_BG}
The observation that fine-tuning greatly diminishing the model compressibility suggests that tweaking of parameters, even if very minor, introduces a lot of entropy. This phenomena is amplified by the use of floating point arithmetic which, by design, invests a significant amount of bits even to very small numbers. However, in reality, parameters with very small numbers tend to have a very minor effect on the results of model inference, if at all. 
This suggests that removing some of the entropy dedicated to very small numbers could increase model compressibility without actually changing the model results or accuracy.  


Thus we introduce a {\em tunable lossy compression} technique that may significantly improve compression savings at the expense of small measurable changes to the original model.
In a nutshell, the proposed method casts every parameter into an integer representation with a chosen level of fixed precision, in essence trimming some of the least bits. Then compression follows as before using byte grouping and a standard lossless compressor. The full decompression of our tunable lossy technique returns the model to its original format albeit zeroing some information that resided in the least bits. 

Formally, given a parameter $\theta$ in floating point representation, and precision $B=2^{b}$ the casting is done as follows.  First, multiply the parameter by the precision factor and then cast it into an integer, effectively rounding it to $\lfloor \theta\cdot B \rfloor$. 
The transformed parameters are then fed into a standard lossless compressor. During decompression, the stream first undergoes standard decompression and then the resulting integers are transformed into floating point after division by the precision factor.
Note that floating point parameters typically lie in the range $[-1,1]$. The meaning of multiplying by the precision factor and rounding is that anything smaller than $1$ after the multiplication is discarded. Thus, the greater the precision factor, the more of the parameter's original entropy is kept. Choosing a precision factor of $2^{b}$ essentially means that we only discard quantities that are smaller than $2^{-b}$. Maintaining a higher precision implies preserving more information or less “lossy” compression.  

A small tweak we add deals with outlier values such as parameters that are outside of the range [-1,1]. These are problematic because when multiplied by the precision factor they may overflow the integer range. In such a case we forgo compression of the layer and mark this layer as non-compressible. Note also that compressing unsigned integers and adding the signed bits separately (as described in Section~\ref{sec:lossless}) turns out to be more significant with tunable lossy compression, so our tests include this optimization as well. 
The improvement from separating the sign bit grows as the precision factor drops -- from $\sim$$3\%$ in the case of $B=2^{27}$ to more than $16\%$ for $B=2^{15}$.

\subsection{Compression vs.\ Accuracy with Tunable Lossy Compression} \label{sec:lossy_res}
One desirable trait of this method is that the amount of information lost is tunable, and determined by the precision factor chosen. 
The main question is then, how to choose this precision factor? 

We offer some rules of thumb that help identify levels that we can push our precision factor to without harming the accuracy of general models. 
We expect that there is an accuracy level of weight information beyond which information is actually redundant for the computation. 
This is basically true by construction, as training mechanisms introduce such errors during training, considering it beyond their precision. 
For example, in order to avoid zero values, Adam's epsilon defaults in PyTorch \cite{pytorch} and Tensorflow \cite{tensorflow2015-whitepaper} are set to $10^{-8}$ and to $10^{-7}$ in Keras \cite{chollet2015keras} deeming anything below this precision irrelevant. 
Under this assumption rounding anything below $10^{-8}$ or $10^{-7}$ is a safe choice and choosing a precision factor of $B = 2^{27}$ or $B=2^{24}$ (respectively) will not insert more noise than what is done naturally in the training process. 

A further argument postulates that Floating Point 32 arithmetic inserts precision error all the time - for example, when adding very small numbers to relatively large numbers, the small numbers may lose their accuracy. For FP32 this happens naturally during computations (such as inference and training) introducing errors of up to $2^{-23}$. Under this rationale, using a precision factor of $B=2^{23}$ is not expected to change overall outcomes more than what arbitrary FP32 operations might do. 

Using $B=2^{23}$ offers significant savings and improves compression ratio significantly for the FP32 models, reducing the compressed model by an additional $20\%$, for example reducing wav2vec compression ratio from $\sim$$85\%$ to $\sim$$68\%$. 

To test this rationale we checked two base models with measurable fine-tuned performance. The first set of fine-tuned models are trained variants of T5 on CNN-DM \cite{nallapati2016abstractive}, XSUM \cite{narayan-etal-2018-dont}, SQUAD \cite{rajpurkar-etal-2016-squad}, asQA \cite{stelmakh-etal-2022-asqa}, wikiAns \cite{wikians} WMT22En-Ru \cite{kocmi2022findings}. Those were evaluated using exact match, Rouge-L \cite{lin-2004-rouge} and sacreBleu\cite{post-2018-call}, but due to the heavy cost of fine-tuning to evaluate, we only evaluated the model under two precision values $B= 2^{24}$ and $B= 2^{19}$. The results were very encouraging and we did not observe a significant change in model performance. The compression for these two configurations was $70\%$ and $56\%$ compared to $85\%$ with only lossless compressors. 

\begin{figure}[ht]
    \begin{minipage}[b]{0.96\linewidth}
        \centering
        \includegraphics[width=\linewidth]{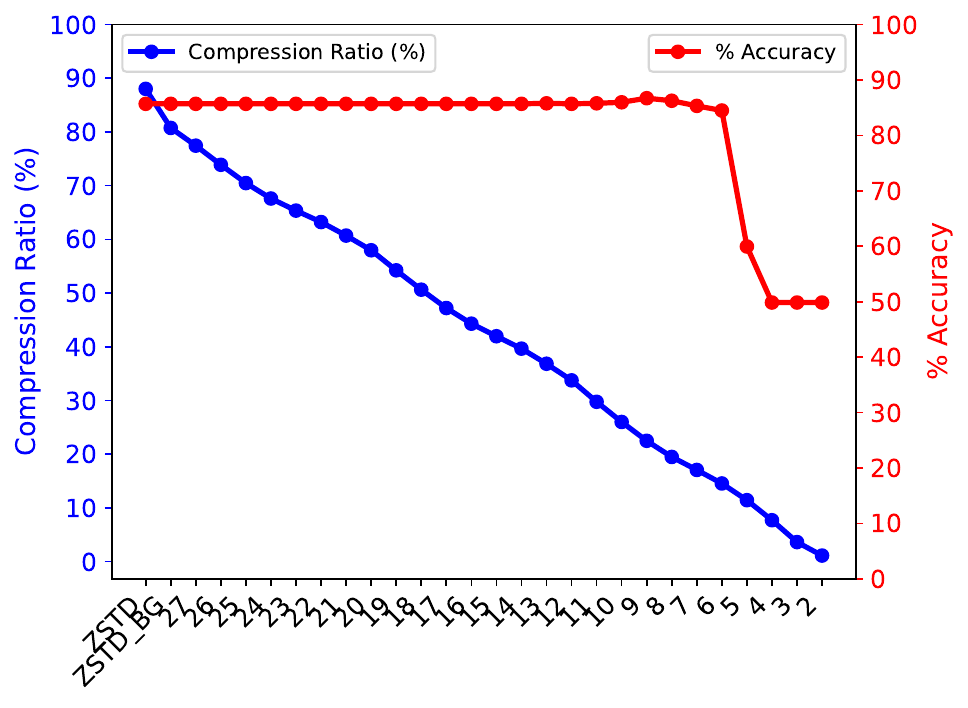}
        \vspace{-0.25in}
       \caption{Fine tuned RoBERTa compression and accuracy as a function of the precision factor parameter b (i.e., for $b=27$ the factor is $B=2^{27}$. The first two values are lossless compression without and with byte grouping. }
        \label{fig:RoBERTa}
    \end{minipage}
\end{figure}

We ran a more detailed test on RoBERTa fine-tuned on the RottenTomatos dataset \cite{rottenTomatos} and the results are shown in Figure~\ref{fig:RoBERTa}.  We see that model accuracy remains high (near $90\%$) until $B=2^6$ and drops dramatically beyond that. The compression on the other hand improves at an almost linear rate culminating at around $20\%$ before the drop-off. Note that there is even a slight rise in accuracy just before the drop, a phenomenon that we saw also in other tests. Note that running this method on the ``clean" version of RoBETRa yields no benefits before reaching $B=2^{18}$ which aligns with our understanding. 

These large cuts in precision with no change in accuracy have empirical and theoretical implications. Empirically, they suggest that much higher compression gains could be achieved with tuning. Theoretically, it may mean there are other training factors stronger than the computation errors above. Possibly, those are not calculation errors but traits of the network such as noise in gradient updates \cite{panigrahi2019non}.

This technique is similar in nature to quantization techniques but is more tunable than quantization with the flexible precision factor, whereas quantization is limited to sizes that can naturally run the actual models like 16 or 8 bits. We elaborate on previous works suggesting low sensitivity to precision in \S\ref{sec:related}.

\section{Evaluation} \label{sec:eval}

\subsection{Integration in PyTorch}\label{sec:integration}
In order to make our compression applicable and easy to use we integrated these methods into PyTorch functions. pytorch.save() and pytorch.load(). These functions are used during uploading and downloading models from Hugging~Face. This integration is the base for our timing evaluation below.  
Our intention is to contribute this code upstream into Pytorch in order to  make compression a standard easy to use component for the community.  

We implemented two approaches to compression, one compressing the entire model as a bin file and the other compressing each layer separately. The latter approach turned out faster for downloads (and slightly slower for uploads), and we use it in the model hub use-case that we evaluate next, which is downloads dominant. 

\remove{
We implemented two approaches to compression:
\begin{itemize}
    \item \textbf{Full Model Compression} - Compresses the full model as one large bin file (ignoring structure). 
    \item \textbf{Per-Layer Compression} -  Compresses each layer separately before writing it to the storage as a bin file. This opens the door to using different compression for different layers, e.g. deciding to skip compression of some layers, or using different precision factor per layer when doing Tunable Lossy Compression.  
\end{itemize}
For various reasons including buffer types and conversion issues, the approaches differ in performance.  
The per-layer compression is faster on the decompression side whereas the full model compression is better on the compression side. Depending on the application one may choose the best approach. In the model hub use-case that we evaluate next, downloads are more dominant and hence we prefer to choose the per-layer model approach. 
}

\subsection{Setup}\label{sec:setup}
In Sections~\ref{sec:losseless_res} and \ref{sec:lossy_res} we present the compressibility traits of various models. In this section, we focus on time aspects and end-2-end timing of our first use-case - that of model hubs. 
We measured the time it takes to upload and download from Hugging~Face to a virtual machine that runs on one of the cloud providers and is Located in the Milan region. We also measured upload and download performance on a home laptop with a 500Mbps network. 

Unlike storage benefits, communication speeds depend heavily on the medium. We first characterized the general behavior of the communication with the Hugging~Face hub. 
The upload bandwidth observed in the cloud remained mostly constant (at around 20 MBps).
On the downloads, we observed 2 types of data transfer speeds. 
\begin{itemize}
    \item \textbf{First Download} - The speed in the first download showed large variance was between 20-40 MBps on the cloud VM. The home machine got approximately 10MBps.
    \item \textbf{Cached Download} - From the second read on the data is likely downloaded from a cloud cache and exhibits speed of 120-130 MBps in the cloud and approximately 40MBps at the home location. 
\end{itemize}

We measured timing with lossless compression on 3 models, one from each of the model groups presented in Section \ref{sec:losseless_res}. Specifically, we used wav2vec, XLM-RoBERTa and Openllama. 
We used ZSTD in default setting (level 3) and byte grouping. We also measured the timing of lossy compression on wav2vec.


\subsection{Results}\label{sec:results}
\begin{figure}[ht]
    \begin{minipage}[b]{0.99\linewidth}
        \centering
        \includegraphics[width=\linewidth]{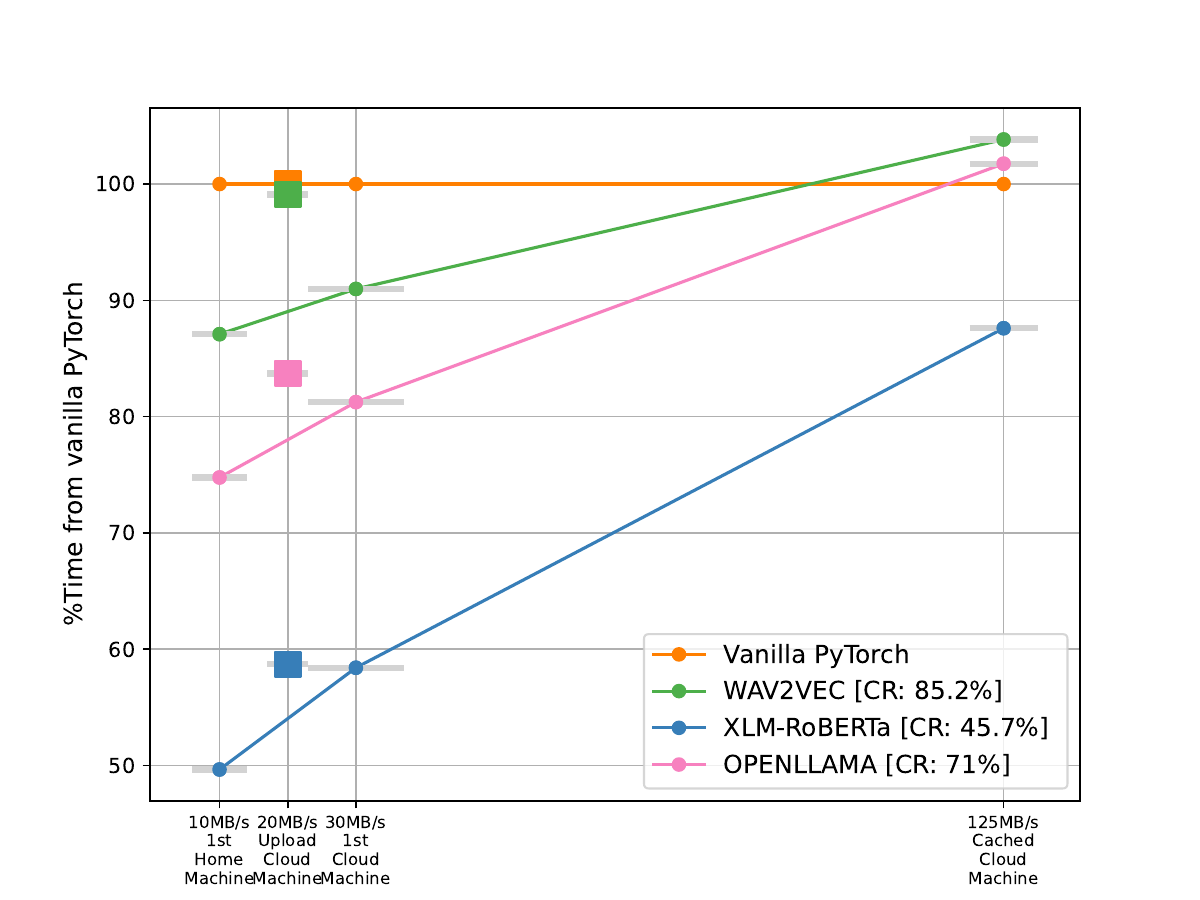}
        \vspace{-0.3in}
        \caption{Download and upload times of 3 models using full model compression vs.\ the non-compressed version. }
        \label{fig:HFtimes}
    \end{minipage}
\end{figure}

The end-2-end timing behavior is dictated by the time to compress/decompress the model and the time to upload/download it respectively. There is additional work done that is mostly constant and does not change with compression. For the overall time to be better than vanilla torch.save and torch.load, the benefits of uploading/downloading less data need to overcome the overhead of the actual compression/decompression. 
Figure~\ref{fig:HFtimes} shows the timing of upload and download of three models. 
Each test was run 10 times for the cached reads and 5 times for the $1^{st}$ timers. The variance was almost entirely due to the network time and this standard deviation is depicted in the graph. The actual compression and decompression time had very little variance. For example, in the xlm-RoBERTa the average time for load part (which includes the decompression) was 3.92 seconds with a standard deviation of 0.017. 

As expected, highly compressible models show significant time improvements whereas the less compressible model category struggles to maintain the same non-compressed timing (but for the most part manages to do so). 
Naturally, the time saving is more significant when the network is slower. Therefore the cached reads in the cloud hardly save time even for the clean model and add a small overhead for the less compressible model. The upload time shows significant time improvement since the bandwidth for uploads is low. On the other hand, the upload savings are lower than download with similar bandwidth reflecting the fact that compression is slower than decompression.  

\begin{figure}[ht]
    \centering
    \begin{minipage}[b]{0.8\linewidth}
        \centering
        \includegraphics[width=\linewidth,  height =1.8in]{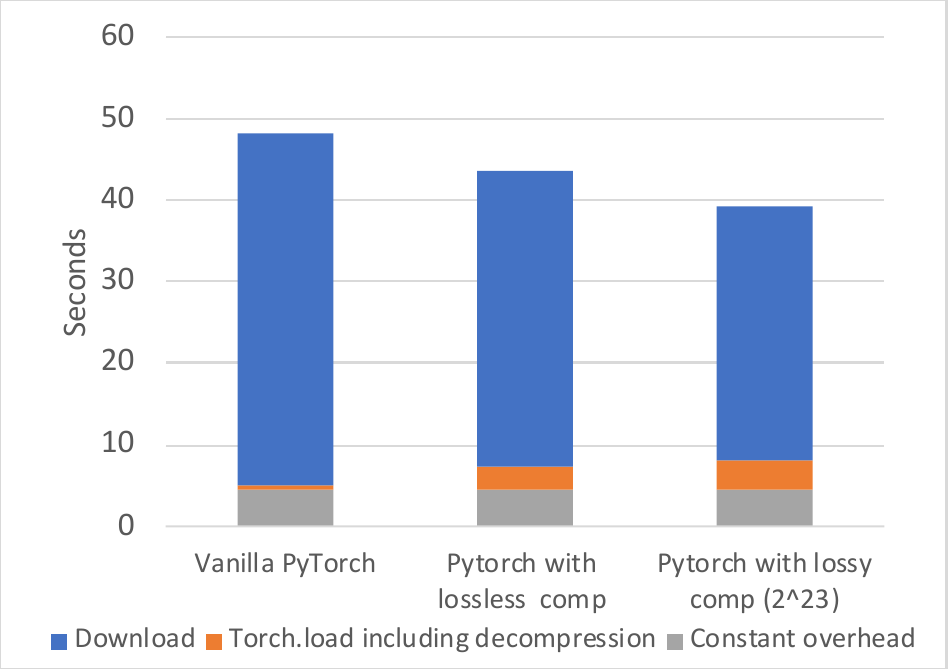}
        \vspace{-0.3in}
        \caption{Breakdown of download time for the wav2vec with a 30MBps network.}
        \label{fig:download_breakdown}
    \end{minipage}
\end{figure}

For the first group of models that are hardly compressed, it is worthwhile to also test the tunable lossy compression. Figure~\ref{fig:download_breakdown} shows a breakdown of the download time of the wav2vec model on a 30MBps network and includes lossy compression with precision parameter $2^{23}$. As expected for this model category the time savings with lossless is marginal, yet with the lossy compression it manages to reduce download time by almost $20\%$ and maintains a slight edge over vanilla Pytorch even with the cloud cache network of 120MBps. 

\begin{figure}[ht]
    \centering
    \begin{minipage}[b]{0.9\linewidth}
        \centering
        \includegraphics[width=\linewidth, height = 1.8in]{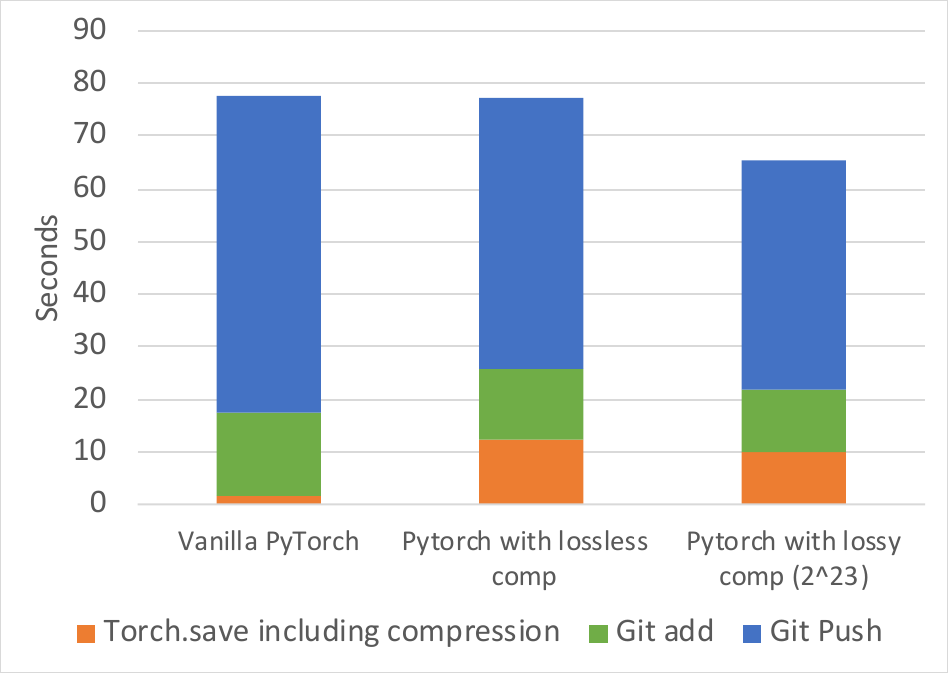}
        \vspace{-0.3in}
        \caption{Breakdown of upload time for the wav2vec model with a 20MBps network.}
        \label{fig:upload_breakdown}
    \end{minipage}
\end{figure}

The breakdown of upload time is presented in Figure~\ref{fig:upload_breakdown}. The lossless technique was just $1\%$ faster than vanilla PyTorch while the lossy version managed to save $16\%$ of the upload time. Note that the tested implementation of tunable lossy did not include the sign bit optimization, as its time overhead did not justify the improved compression ratio. The lossy compression ended up with a $72\%$ compression ratio (versus $85\%$ with the lossless compression). 
Interestingly, the compression time of lossy compression was slightly faster than in the lossless case. This is likely due to the improved compression ratio which typically translates to faster compression speeds. 

\remove{
\begin{table*}[t]
\caption{Saving percentage of Models after lossless compression [The models were downloaded on August 2023]}
\label{table:hf_precent}
\vskip 0.15in
\begin{center}
\begin{small}
\begin{sc}
\begin{tabular}{lcccccr}
\toprule
Model &  Compression Type  & upload & Download From cache &  Download First Time \\ 
Name & Type & (save(), git add, git push) & (download, load())& (download, load())\\ 

Wav2vec & NONE & 100\% () & 100\% () & 100\% () \\
(Group A) & OP\_A ZSTD & 97.8\% & 127\% &  \\
& OP\_A BG\_ZSTD & \textbf{91.8\%} & \textbf{121\%} & \\
& OP\_B ZSTD & 101.15\% & 119\% & \\
& OP\_B BG\_ZSTD & \textbf{103.5\%} & \textbf{103.6\%} & \textbf{92\%-...}\\ \hline

XLM-RoBERTa & NONE & 100\% (1.3, 14.2, 63.5) & 100\%  &  100\% () \\
(Group B) & OP\_A ZSTD & 56.2\% (4.6, 8.3. 31.5) & 96\% & \\
& OP\_A BG\_ZSTD & \textbf{43.3\%} (4.7, 6.4, 23.1) & \textbf{94\%} & \\
& OP\_B ZSTD & 54.38\% (8.0, 8.0, 27) & 90.7\% & \\
& OP\_B BG\_ZSTD & \textbf{49.05\%} (9.5, 6.3, 23) & \textbf{85.1\%} & \\ \hline

Openllama & NONE & 100\% () & 100\% () & 100\% () \\
(Group C) & OP\_A ZSTD & 77.1\% & 112.0\% & 82\% \\
& OP\_A BG\_ZSTD & \textbf{61.4\%} & \textbf{111.49} & \textbf{... - 82\%} \\
& OP\_B ZSTD & 79.8\% & 106.2\% & 269\danny{error?} \\
& OP\_B BG\_ZSTD & \textbf{63.4\%} & \textbf{99.5\%} & \textbf{90\%-82\%} \\ 
\bottomrule
\end{tabular}
\end{sc}
\end{small}
\end{center}
\end{table*}
}

\section{Beyond Full Model Compression}\label{sec:beyond}
 
\subsection{Delta compression.}\label{sec:deltas} 
When models have high similarity, one strategy to optimize storage and network transfer is to save a base model and for the rest of the models only store the differences from this base model \cite{kandpal2023git}. We refer to compressing those differences as \emph{delta compression}. To reconstruct a model, one only needs to apply the delta to the base model. A straightforward approach to delta compression is to compute the difference between the two models (e.g. using XOR or subtraction) and compress this delta using a standard compressor. 

\begin{figure*}[ht]
    \begin{minipage}[b]{0.48\linewidth}
        \centering
        \includegraphics[width=\linewidth]{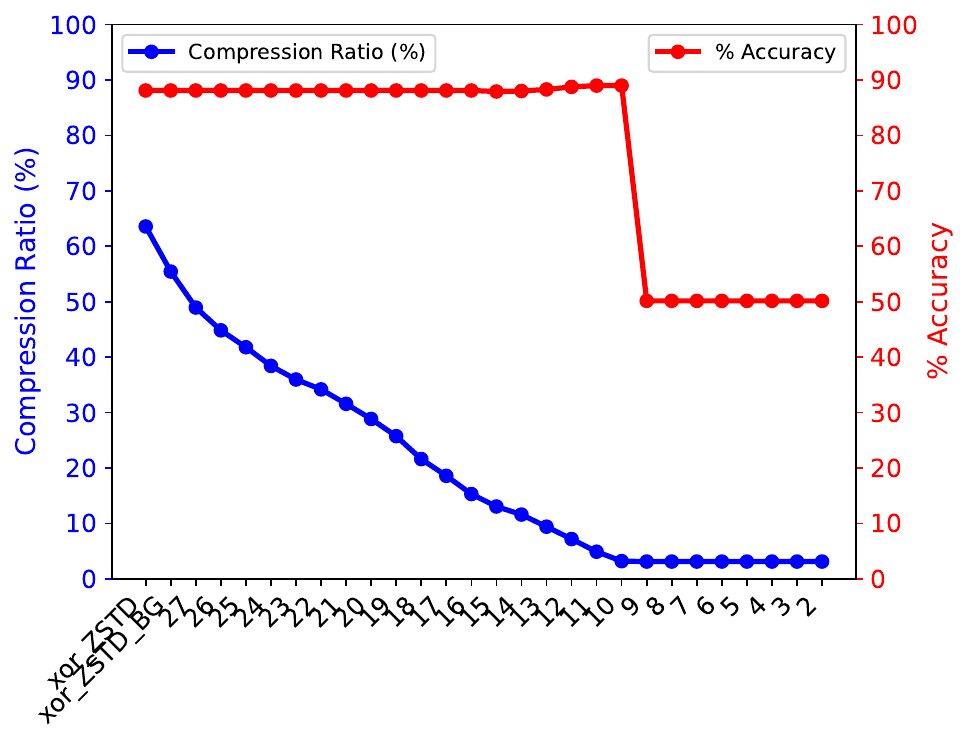}
        \caption{Compression ratios and accuracy of the delta between two fine tuned RoBERTa model in consecutive epochs (after 10 epochs and after 9 epochs). }
        \label{fig:delta_base}
    \end{minipage}
    \hfill 
    \begin{minipage}[b]{0.48\linewidth}
        \centering
        \includegraphics[width=\linewidth]{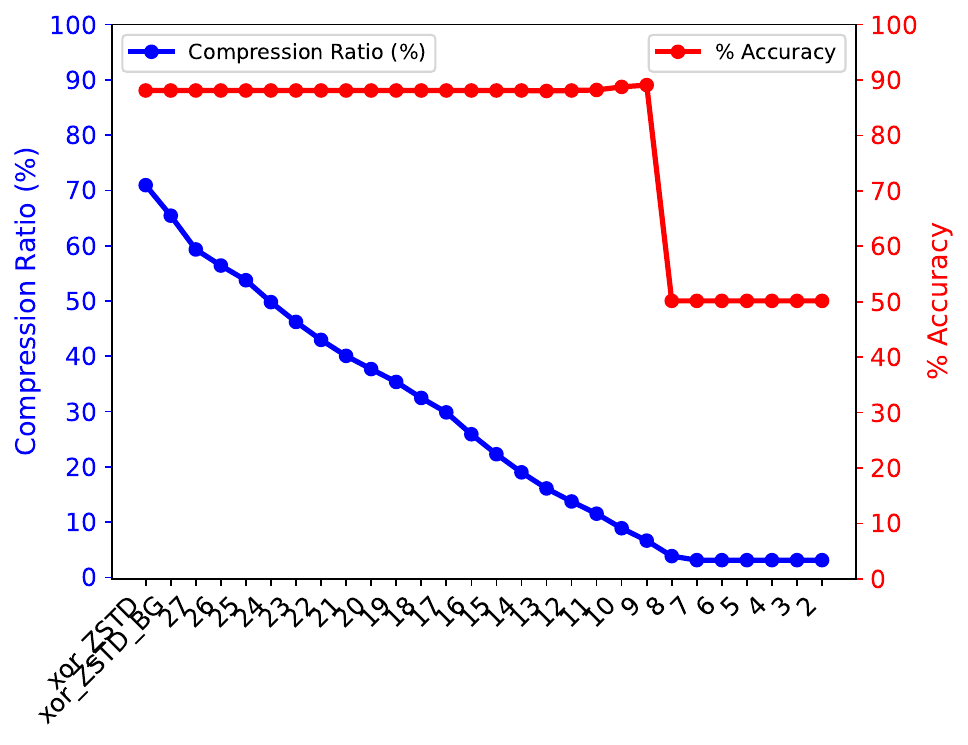}
        \caption{Compression ratios and accuracy of the delta between the fine tuned RoBERTa model after 10 epochs and the base. }
        \label{fig:delta_consec}
    \end{minipage}
\end{figure*}

\begin{figure}[ht]
    \begin{minipage}[b]{0.96\linewidth}
        \centering
        \includegraphics[width=\linewidth]{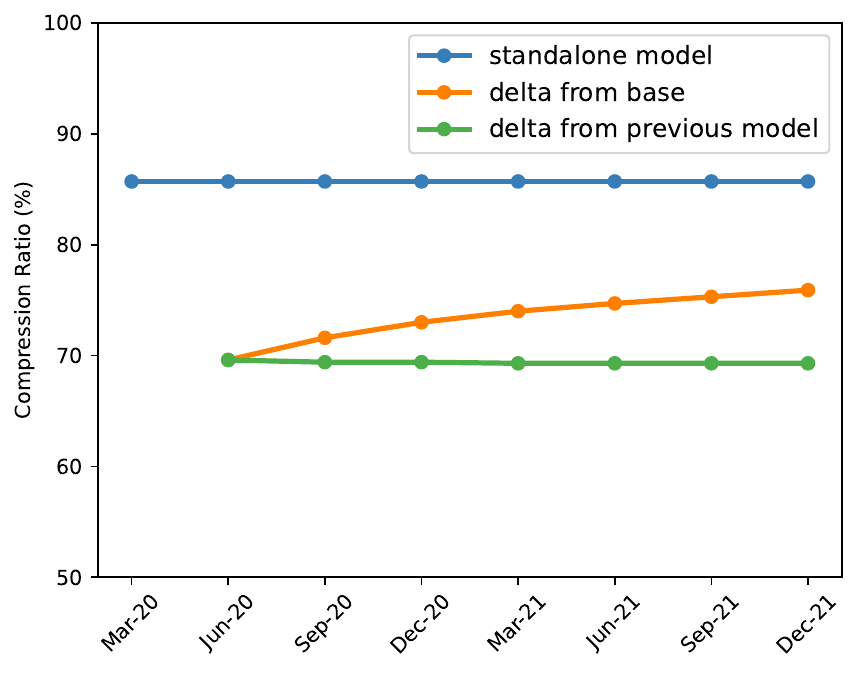}
        \caption{Compression ratio (lossless with byte grouping) of RoBERTa finetuned on tweets that occurred up until the month in the X-axis. We compare the compression of the standalone model to compression of deltas of consecutive models and compression of delta from one base.}
        \label{fig:finetune}
         \end{minipage}
\end{figure}

A natural use case in which delta compression proves very useful is checkpointing. 
In checkpointing, we repeatedly store models that have limited change between them. As mentioned in discussing tunable lossy compression (\S\ref{sec:lossy_BG}), fine tuning often changes models by small quantities. Hence the delta between the results of consecutive training epochs is highly compressible. This is true for lossless compression (especially with byte grouping) as well as tunable lossy compression. Figure~\ref{fig:delta_consec} shows this for consecutive fine tuned epochs of the RoBERTa model. We see that lossless compression is as low as 55\% (with byte grouping), down from nearly 83\% of this model standalone. Figure~\ref{fig:delta_base} shows the compressibility of the $10^{th}$ epoch vs.\ the base RoBERTa which is useful as it avoids maintaining long chains of deltas. In this case, the delta is less compressible but still achieves a 65\% compression ratio. Using tunable lossy compression proves to be very beneficial also in delta compression. For example, taking $B=2^{23}$ achieves a compression ratio of 37\% for consecutive models and 49\% vs.\ the base model without affecting model accuracy. 
It is worth noting that an aggressive choice of the precision factor can achieve below 10\% (over 90\% savings!) without harming accuracy and in fact even achieving a slight improvement to the accuracy.

Another use-case is when a hub or user stores multiple models with high similarity (regardless of checkpointing). One source for such occurrence is when multiple models are trained or fine tuned from the same base model. For example, we found in Hugging~Face 3 variations of RoBERTa trained on tweets and fine tuned for different purposes (For the exact model names, see Appendix~\ref{app:deltas_models}) - detecting irony, detecting offensive language and detecting abuse. As standalone models, their compression ratio (lossless with byte grouping) is 85.7\% on average. However, when compressing the delta of each of the pairs achieves a ratio of 56\% on average. 

Another example is a set of models that are fine tuned on the twitter data set once every 3 months. 
The results, shown in Figure~\ref{fig:finetune}, show that delta compression is most beneficial in consecutive versions and its effectiveness slowly deteriorates over time.


\begin{figure}[ht]
    \begin{minipage}[b]{0.96\linewidth}
        \centering
        \includegraphics[width=\linewidth]{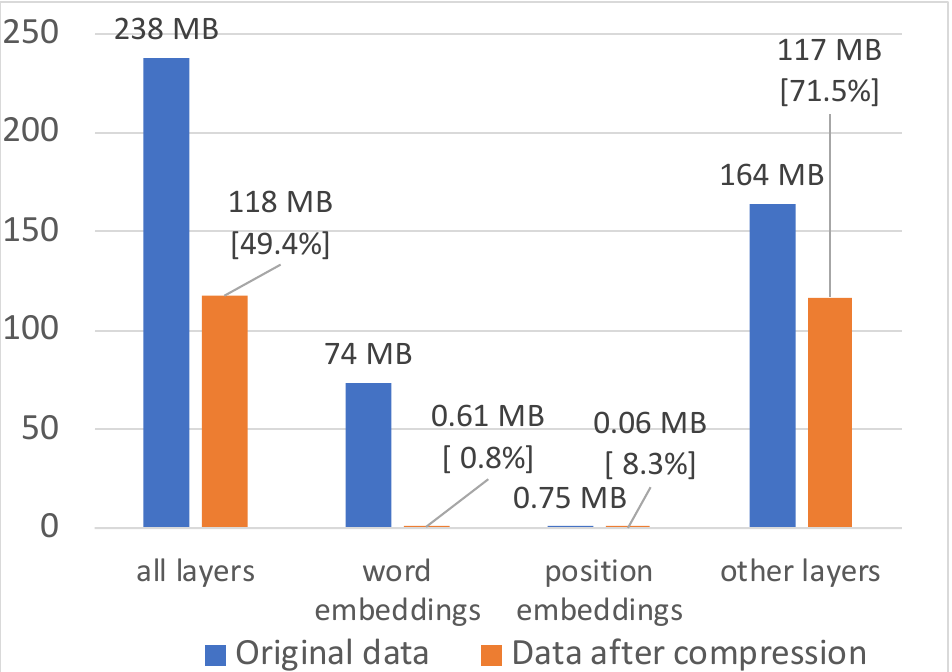}
       \vspace{-0.25in}
        \caption{Compressibility of layers in the {\bf Gradients}}
        \label{fig:gradients}
    \end{minipage}
\end{figure}
\begin{figure}[ht]
    \begin{minipage}[b]{0.96\linewidth}
        \centering
        \includegraphics[width=\linewidth]{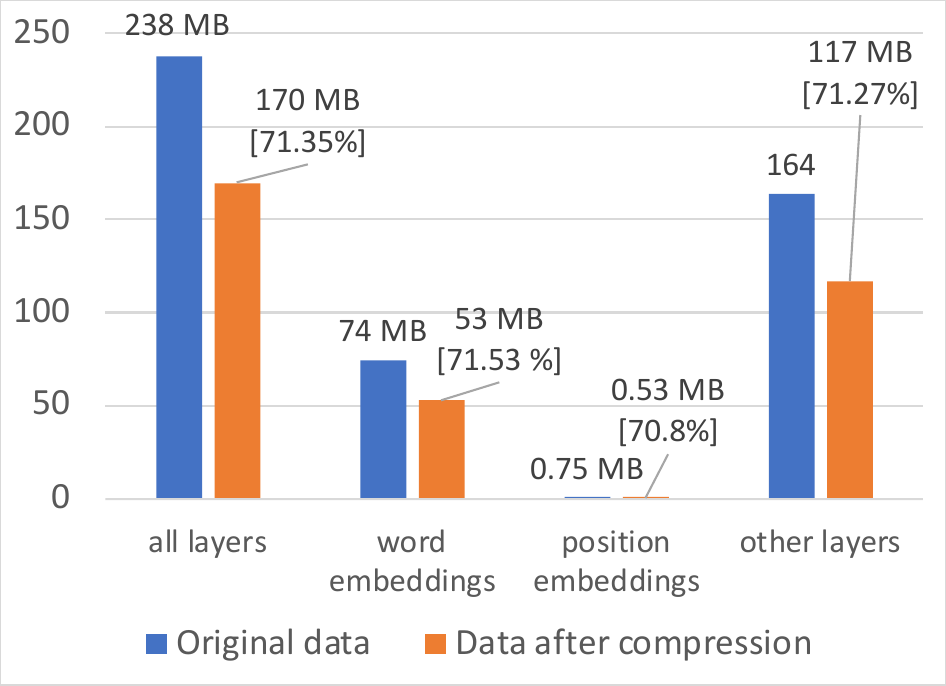}
       \vspace{-0.25in}
        \caption{Compressibility of layers in the {\bf model}}
        \label{fig:model}
    \end{minipage}
\end{figure}

\begin{figure}[ht]
    \begin{minipage}[b]{0.96\linewidth}
        \centering
        \includegraphics[width=\linewidth]{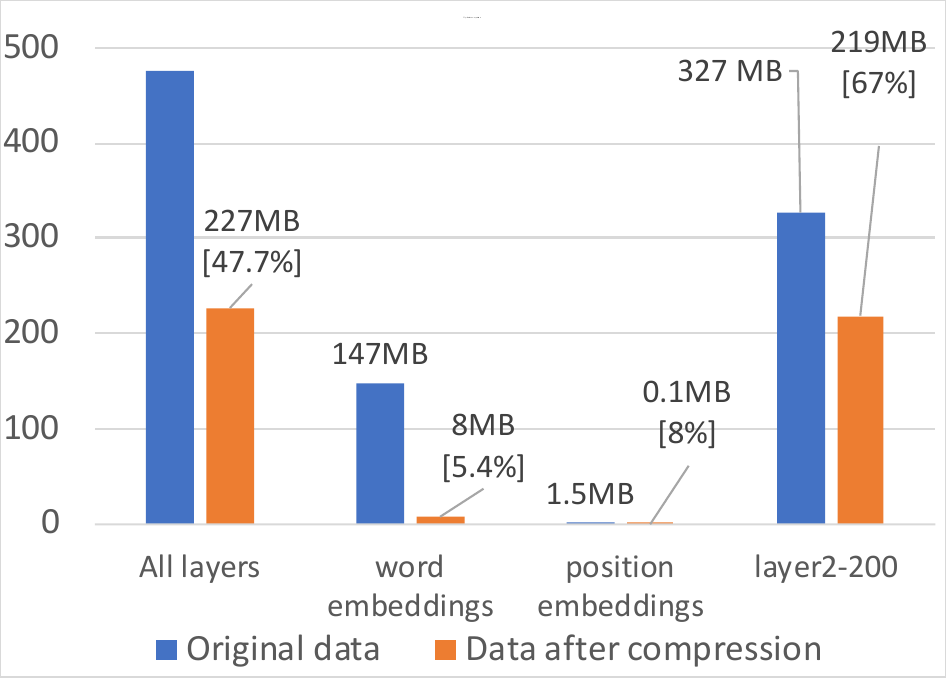}
       \vspace{-0.25in}
        \caption{Compressibility of layers in the {\bf Optimizer}}
        \label{fig:optimizer}
    \end{minipage}
\end{figure}

\subsection{Compressing Gradients and Optimizers}\label{sec:gradients}
So far, we have mostly focused on compressing the actual models, but in various cases, such as distributed training or checkpointing, derivatives of the model also take up resources. Specifically, Gradients and Optimizers usually occupy a substantial amount of the communication, often of equal in size to the models \cite{anil2020scalable}. 
We find that these components can also benefit from compression. 

We investigate a BF16 version of RoBERTa. Figure~\ref{fig:gradients} shows the break down of compressibility of the gradients broken down according to different layers. This is contrasted with the breakdown of the actual model in Figure~\ref{fig:model}. Not only that we find that most of the weights in the gradient compress similarly to regular weights. We find that token embeddings are extremely compressible, despite not showing a different behavior in regular models. 

The optimizers show similar compressibility traits to the gradients. Namely, the embedding layer is extremely compressible and the general layers compress to around $67\%$, slightly better than these layers in the model itself as shown in Figure~\ref{fig:optimizer}.

\begin{table*}[t]  
\caption{Compression Ratio of quantized models after Zstd compression with Byte Grouping.}
\label{table:quantization}
\vskip 0.15in
\begin{center}
\begin{small}
\begin{sc}
\begin{tabular}{lcccccr}
\toprule
Model & Param & Model & Compression  &  Compression Ratio\\
name & Type & Size &  Ratio & Per Byte Group  \\
\hline \hline
\textbf{CapybaraHermes-Mistral} & FP16 & 14.5 GB &  \textbf{83.7\%} & (68.8\%, 98.7\%) 
\\
\hline
\textbf{CapybaraHermes-Mistral GPTQ 8b 1GB actorder} & 8bit & 7.5 GB &  \textbf{86.6\%} &  (86.5\%, 86.6\%)
\\
\textbf{CapybaraHermes-Mistral GPTQ 8b 32GB actorder} & 8bit & 8.2 GB &  \textbf{90.3\%} &  (88.5\%, 92.1\%)
\\
\textbf{CapybaraHermes-Mistral GPTQ 8b 128GB actorder} & 8bit & 7.7 GB &  \textbf{91.2\%} &  (90.8\%, 91.7\%)
\\
\hline
\textbf{CapybaraHermes-Mistral GPTQ 4b 32GB actorder} & 4bit & 4.6 GB &  \textbf{85.4\%} &  (82\%, 88\%)
\\
\textbf{CapybaraHermes-Mistral GPTQ 4b 64GB actorder} & 4bit & 4.3 GB & \textbf{84.7\%} &  (83.2\%, 86.25\%)
\\
\hline
\textbf{CapybaraHermes-Mistral AWQ 4b} & 4bit & 4.15 GB &  \textbf{87.6\%} &  (86.9\%, 88.3\%)
\\
\bottomrule
\end{tabular}
\end{sc}
\end{small}
\end{center}
\end{table*}

\section {Related Work} \label{sec:related}
\subsection{Model Compression}
In the literature, "model-compression" is a field of its own, aimed at creating smaller models, that mimic the original model. Model-compression is hence the name for a set of tools that aim to \emph{accelerate} models, usually at inference, by reducing their size \cite{choudhary2020comprehensive}. Under such conditions, a method is allowed to reduce the accuracy, and is judged on its tradeoff between size and performance. This differs from lossless compression which is supposed to return the model to its original state after decompression. 

There are four main methods to reduce model size in that manner \cite{choudhary2020comprehensive}. Pruning \cite{LeCun1989OptimalBD,Hanson1988ComparingBF,zhu2017prune} (sometimes referred to as sparsification;  \cite{ma2021effective}) where parts of the model are removed, dedicated training or network architecture \cite{oktay2019scalable}, distillation \cite{gou2021knowledge} or otherwise training a smaller model from a better model \cite{Haroush2019TheKW} and quantization \cite{gholami2021survey_quantization}. There are also methods combining several of those \cite{polino2018model}, including the only work we have found to propose compression, which it applies after two other model-compression steps \cite{Han2015DeepCC}.

\subsection{Quantization}
Out of the model-compression techniques, quantization is the most similar to the tunably lossy compression we discuss in this paper (\S\ref{sec:lossy_BG}). Quantization \cite{gray1984vector} is a method that bins weight values to a more coarse granularity. 
Since the model would be used as is, quantization is limited in the granularity to which it can compress models. For example, it cannot reduce to 23 bits (typically 16 is the first viable choice). Moreover, quanitization is not optimized for the smallest representation of the model and in fact quantized models can potentially be further compressed. We examine off-the-shelf quantized models (For the exact model names,  See Appendix~\ref{app:quantized}) that have been quantized with GTPQ~\cite{frantar2022gptq} and AWQ~\cite{lin2023awq} and compress them using lossless compression. We see in Table~\ref{table:quantization} that they are still compressible, with a compression ratio between 85-91\%, where byte grouping contributes to the compression 1-2\%.

In a sense, quantization is complementary to compression. Quantization is able to improve inference speed and to reduce model size  drastically, at costs to inference accuracy or the ability to further train the model. In contrast, compression cannot speed inference but can compress models further, or compress a model without affecting its behaviour at all. Similar to quantization, our Tunable lossy compression drops some of the information, but only ones that don't reduce accuracy in a measurable way. Unlike quantization, it changes the representation which is expected to further reduce the size as seen by the results above. 



\subsection{Other Related Work}
Interestingly, two recent works also found improved results by reducing some of the information in the network (see \S\ref{sec:losseless_res}). They saw it during pruning \cite{Sharma2023TheTI} or pruning and extreme quantization \cite{yadav2023compeft}. We observe this same phenomena in the tunable lossy compression. Future work may find a theory to connect those findings.

Similar to delta compression some works analyze dimensionality \cite{aghajanyan-etal-2021-intrinsic, gueta-etal-2023-knowledge} or save  the deltas for actions such as compositionally \cite{ilharco2022editing} and merging multiple deltas \cite{Choshen2022FusingFM,Wortsman2022ModelSA,matena2021merging}. Few works also apply on such deltas the above methods, like pruning \cite{yadav2023ties}, trained sparsity \cite{zhang2023adaptive} or quantization \cite{dettmers2023qlora} or discuss deltas.

Another line of work worth mentioning is computation graph optimization \cite{sabne2020xla,wu2023pytorch}. Such works reduce the computation graph and perform optimization there. It is mostly noteworthy to contrast it to our work, such work compresses the size of the computation graph, which speeds computation, but does not change the model weights, and it is hence orthogonal to our work. To validate that, we compressed pyTorch \cite{wu2023pytorch} models before and after compilation, finding compression works similarly well.

Last, related to the optimization process (see \S\ref{sec:gradients}), a few recent works offer to reduce the information passed in gradient updates hence making them faster, faster overcoming the reduce in information per example by seeing more examples per second \cite{tyagi2023gravac,zhao2024galore}.

\section {Conclusion}
We are in an era where models and system requirements grow larger,  overparametrization seems to be beneficial for better learning. As our compression findings hint, this overparametrization is not fully used for inference or for the weights themselves and there is redundancy. Hence the wide attention and progress made to reducing model sizes is not without merit. 
That being said, the reality is that commonly used models are not kept or run in reduced form and there is great inefficiency in the way models are stored and communicated today. Some of this inefficiency can be mitigated using the compression techniques outlined in this paper. 

Given the reduction in network bandwidth, storage and time, we think that lossless compression should be the default in communication with model hubs such as Hugging~Face. Moreover, we believe that communication compression has multiple other use-cases in the realm of training, versioning and serving models. 


\bibliography{bib,example_paper}

\begin{thebibliography}{10}

\bibitem{Biderman2023PythiaAS}
S.~Biderman, H.~Schoelkopf, Q.~G. Anthony, H.~Bradley, K.~O'Brien, E.~Hallahan, M.~A. Khan, S.~Purohit, U.~S. Prashanth, E.~Raff, A.~Skowron, L.~Sutawika, and O.~van~der Wal, ``Pythia: A suite for analyzing large language models across training and scaling,'' {\em ArXiv}, vol.~abs/2304.01373, 2023.

\bibitem{kandpal2023git}
N.~Kandpal, B.~Lester, M.~Muqeeth, A.~Mascarenhas, M.~Evans, V.~Baskaran, T.~Huang, H.~Liu, and C.~Raffel, ``Git-theta: A git extension for collaborative development of machine learning models,'' {\em arXiv preprint arXiv:2306.04529}, 2023.

\bibitem{don-yehiya-etal-2023-cold}
S.~Don-Yehiya, E.~Venezian, C.~Raffel, N.~Slonim, and L.~Choshen, ``{C}ol{D} fusion: Collaborative descent for distributed multitask finetuning,'' in {\em Proceedings of the 61st Annual Meeting of the Association for Computational Linguistics (Volume 1: Long Papers)} (A.~Rogers, J.~Boyd-Graber, and N.~Okazaki, eds.), (Toronto, Canada), pp.~788--806, Association for Computational Linguistics, July 2023.

\bibitem{zhang2021survey}
C.~Zhang, Y.~Xie, H.~Bai, B.~Yu, W.~Li, and Y.~Gao, ``A survey on federated learning,'' {\em Knowledge-Based Systems}, vol.~216, p.~106775, 2021.

\bibitem{Wolf2019HuggingFacesTS}
T.~Wolf, L.~Debut, V.~Sanh, J.~Chaumond, C.~Delangue, A.~Moi, P.~Cistac, T.~Rault, R.~Louf, M.~Funtowicz, and J.~Brew, ``Huggingface's transformers: State-of-the-art natural language processing,'' {\em ArXiv}, vol.~abs/1910.03771, 2019.

\bibitem{jiang2023mistral}
A.~Q. Jiang, A.~Sablayrolles, A.~Mensch, C.~Bamford, D.~S. Chaplot, D.~d.~l. Casas, F.~Bressand, G.~Lengyel, G.~Lample, L.~Saulnier, {\em et~al.}, ``Mistral 7b,'' {\em arXiv preprint arXiv:2310.06825}, 2023.

\bibitem{gou2021knowledge}
J.~Gou, B.~Yu, S.~J. Maybank, and D.~Tao, ``Knowledge distillation: A survey,'' {\em International Journal of Computer Vision}, vol.~129, pp.~1789--1819, 2021.

\bibitem{ma2021effective}
X.~Ma, M.~Qin, F.~Sun, Z.~Hou, K.~Yuan, Y.~Xu, Y.~Wang, Y.-K. Chen, R.~Jin, and Y.~Xie, ``Effective model sparsification by scheduled grow-and-prune methods,'' {\em arXiv preprint arXiv:2106.09857}, 2021.

\bibitem{gholami2021survey_quantization}
A.~Gholami, S.~Kim, Z.~Dong, Z.~Yao, M.~W. Mahoney, and K.~Keutzer, ``A survey of quantization methods for efficient neural network inference,'' 2021.

\bibitem{deutsch1996zlib}
P.~Deutsch and J.-L. Gailly, ``Zlib compressed data format specification version 3.3,'' tech. rep., 1996.

\bibitem{collet2018zstandard}
Y.~Collet and M.~Kucherawy, ``Zstandard compression and the application/zstd media type,'' tech. rep., 2018.

\bibitem{Devlin2019BERTPO}
J.~Devlin, M.-W. Chang, K.~Lee, and K.~Toutanova, ``Bert: Pre-training of deep bidirectional transformers for language understanding,'' in {\em North American Chapter of the Association for Computational Linguistics}, 2019.

\bibitem{Fedus2021SwitchTS}
W.~Fedus, B.~Zoph, and N.~M. Shazeer, ``Switch transformers: Scaling to trillion parameter models with simple and efficient sparsity,'' {\em J. Mach. Learn. Res.}, vol.~23, pp.~120:1--120:39, 2021.

\bibitem{modelZoo}
J.~Yu~Koh, ``Model zoo (hub),'' 2018.

\bibitem{Pytorch_2019}
Pytorch, ``Pytorch hub,'' 2019.

\bibitem{tensorflowHub}
Google, ``Tensorflow hub,'' 2018.

\bibitem{adapterhub}
J.~Pfeiffer, A.~R{\"u}ckl{\'e}, C.~Poth, A.~Kamath, I.~Vuli{\'c}, S.~Ruder, K.~Cho, and I.~Gurevych, ``{A}dapter{H}ub: A framework for adapting transformers,'' in {\em Proceedings of the 2020 Conference on Empirical Methods in Natural Language Processing: System Demonstrations} (Q.~Liu and D.~Schlangen, eds.), (Online), pp.~46--54, Association for Computational Linguistics, Oct. 2020.

\bibitem{watsonxdata}
``Ibm watsonx.data.'' https://www.ibm.com/products/watsonx-data.

\bibitem{QualcommAIHUB}
``Qualcomm® ai hub.'' https://aihub.qualcomm.com/.

\bibitem{zhao2023pytorch}
Y.~Zhao, A.~Gu, R.~Varma, L.~Luo, C.-C. Huang, M.~Xu, L.~Wright, H.~Shojanazeri, M.~Ott, S.~Shleifer, {\em et~al.}, ``Pytorch fsdp: experiences on scaling fully sharded data parallel,'' {\em arXiv preprint arXiv:2304.11277}, 2023.

\bibitem{li2022branch}
M.~Li, S.~Gururangan, T.~Dettmers, M.~Lewis, T.~Althoff, N.~A. Smith, and L.~Zettlemoyer, ``Branch-train-merge: Embarrassingly parallel training of expert language models,'' {\em arXiv preprint arXiv:2208.03306}, 2022.

\bibitem{lialin2023relora}
V.~Lialin, S.~Muckatira, N.~Shivagunde, and A.~Rumshisky, ``Relora: High-rank training through low-rank updates,'' in {\em Workshop on Advancing Neural Network Training: Computational Efficiency, Scalability, and Resource Optimization (WANT@ NeurIPS 2023)}, 2023.

\bibitem{NEURIPS2021_41a60377}
M.~Diskin, A.~Bukhtiyarov, M.~Ryabinin, L.~Saulnier, q.~lhoest, A.~Sinitsin, D.~Popov, D.~V. Pyrkin, M.~Kashirin, A.~Borzunov, A.~Villanova~del Moral, D.~Mazur, I.~Kobelev, Y.~Jernite, T.~Wolf, and G.~Pekhimenko, ``Distributed deep learning in open collaborations,'' in {\em Advances in Neural Information Processing Systems} (M.~Ranzato, A.~Beygelzimer, Y.~Dauphin, P.~Liang, and J.~W. Vaughan, eds.), vol.~34, pp.~7879--7897, Curran Associates, Inc., 2021.

\bibitem{Turner2021BayesianOI}
R.~Turner, D.~Eriksson, M.~J. McCourt, J.~Kiili, E.~Laaksonen, Z.~Xu, and I.~M. Guyon, ``Bayesian optimization is superior to random search for machine learning hyperparameter tuning: Analysis of the black-box optimization challenge 2020,'' in {\em Neural Information Processing Systems}, 2021.

\bibitem{Dodge2020FineTuningPL}
J.~Dodge, G.~Ilharco, R.~Schwartz, A.~Farhadi, H.~Hajishirzi, and N.~A. Smith, ``Fine-tuning pretrained language models: Weight initializations, data orders, and early stopping,'' {\em ArXiv}, vol.~abs/2002.06305, 2020.

\bibitem{JunczysDowmunt2018MarianFN}
M.~Junczys-Dowmunt, R.~Grundkiewicz, T.~Dwojak, H.~T. Hoang, K.~Heafield, T.~Neckermann, F.~Seide, U.~Germann, A.~F. Aji, N.~Bogoychev, A.~F.~T. Martins, and A.~Birch, ``Marian: Fast neural machine translation in c++,'' in {\em Annual Meeting of the Association for Computational Linguistics}, 2018.

\bibitem{Sandler2023TrainingTM}
M.~Sandler, A.~Zhmoginov, M.~Vladymyrov, and N.~Miller, ``Training trajectories, mini-batch losses and the curious role of the learning rate,'' {\em ArXiv}, vol.~abs/2301.02312, 2023.

\bibitem{liu2023llm360}
Z.~Liu, A.~Qiao, W.~Neiswanger, H.~Wang, B.~Tan, T.~Tao, J.~Li, Y.~Wang, S.~Sun, O.~Pangarkar, R.~Fan, Y.~Gu, V.~Miller, Y.~Zhuang, G.~He, H.~Li, F.~Koto, L.~Tang, N.~Ranjan, Z.~Shen, X.~Ren, R.~Iriondo, C.~Mu, Z.~Hu, M.~Schulze, P.~Nakov, T.~Baldwin, and E.~P. Xing, ``Llm360: Towards fully transparent open-source llms,'' {\em arXiv}, 2023.

\bibitem{wang2019bfloat16}
S.~Wang and P.~Kanwar, ``Bfloat16: The secret to high performance on cloud tpus,'' {\em Google Cloud Blog}, vol.~4, 2019.

\bibitem{squash}
Squash, ``Squash compression benchmark,'' 2016.

\bibitem{LZ77}
J.~Ziv and A.~Lempel, ``A universal algorithm for sequential data compression,'' {\em IEEE Transactions on Information Theory}, vol.~23, no.~3, pp.~337--343, 1977.

\bibitem{Huffman1952}
D.~A. Huffman, ``A method for the construction of minimum-redundancy codes,'' {\em Proceedings of the IRE}, vol.~40, no.~9, pp.~1098--1101, 1952.

\bibitem{Riss1976AC}
J.~J. Rissanen, ``Generalized kraft inequality and arithmetic coding,'' {\em IBM Journal of Research and Development}, vol.~20, no.~3, pp.~198--203, 1976.

\bibitem{lz4}
Y.~Collet, ``Lz4 - extremely fast compression,'' 2024.

\bibitem{zlib}
M.~Adler and J.-L. Gailly, ``Zlib,'' 2024.

\bibitem{zstd}
Y.~Collet, ``Zstandard,'' 2024.

\bibitem{rottenTomatos}
B.~Pang and L.~Lee, ``Seeing stars: Exploiting class relationships for sentiment categorization with respect to rating scales,'' in {\em Proceedings of the ACL}, 2005.

\bibitem{raffel2020exploring}
C.~Raffel, N.~Shazeer, A.~Roberts, K.~Lee, S.~Narang, M.~Matena, Y.~Zhou, W.~Li, and P.~J. Liu, ``Exploring the limits of transfer learning with a unified text-to-text transformer,'' {\em The Journal of Machine Learning Research}, vol.~21, no.~1, pp.~5485--5551, 2020.

\bibitem{pytorch}
A.~Paszke, S.~Gross, F.~Massa, A.~Lerer, J.~Bradbury, G.~Chanan, T.~Killeen, Z.~Lin, N.~Gimelshein, L.~Antiga, A.~Desmaison, A.~Kopf, E.~Yang, Z.~DeVito, M.~Raison, A.~Tejani, S.~Chilamkurthy, B.~Steiner, L.~Fang, J.~Bai, and S.~Chintala, ``Pytorch: An imperative style, high-performance deep learning library,'' in {\em Advances in Neural Information Processing Systems 32}, pp.~8024--8035, Curran Associates, Inc., 2019.

\bibitem{tensorflow2015-whitepaper}
M.~Abadi, A.~Agarwal, P.~Barham, E.~Brevdo, Z.~Chen, C.~Citro, G.~S. Corrado, A.~Davis, J.~Dean, M.~Devin, S.~Ghemawat, I.~Goodfellow, A.~Harp, G.~Irving, M.~Isard, Y.~Jia, R.~Jozefowicz, L.~Kaiser, M.~Kudlur, J.~Levenberg, D.~Man\'{e}, R.~Monga, S.~Moore, D.~Murray, C.~Olah, M.~Schuster, J.~Shlens, B.~Steiner, I.~Sutskever, K.~Talwar, P.~Tucker, V.~Vanhoucke, V.~Vasudevan, F.~Vi\'{e}gas, O.~Vinyals, P.~Warden, M.~Wattenberg, M.~Wicke, Y.~Yu, and X.~Zheng, ``{TensorFlow}: Large-scale machine learning on heterogeneous systems,'' 2015.
\newblock Software available from tensorflow.org.

\bibitem{chollet2015keras}
F.~Chollet {\em et~al.}, ``Keras.'' \url{https://keras.io}, 2015.

\bibitem{nallapati2016abstractive}
R.~Nallapati, B.~Zhou, C.~Gulcehre, B.~Xiang, {\em et~al.}, ``Abstractive text summarization using sequence-to-sequence rnns and beyond,'' {\em arXiv preprint arXiv:1602.06023}, 2016.

\bibitem{narayan-etal-2018-dont}
S.~Narayan, S.~B. Cohen, and M.~Lapata, ``Don{'}t give me the details, just the summary! topic-aware convolutional neural networks for extreme summarization,'' in {\em Proceedings of the 2018 Conference on Empirical Methods in Natural Language Processing} (E.~Riloff, D.~Chiang, J.~Hockenmaier, and J.~Tsujii, eds.), (Brussels, Belgium), pp.~1797--1807, Association for Computational Linguistics, Oct.-Nov. 2018.

\bibitem{rajpurkar-etal-2016-squad}
P.~Rajpurkar, J.~Zhang, K.~Lopyrev, and P.~Liang, ``{SQ}u{AD}: 100,000+ questions for machine comprehension of text,'' in {\em Proceedings of the 2016 Conference on Empirical Methods in Natural Language Processing} (J.~Su, K.~Duh, and X.~Carreras, eds.), (Austin, Texas), pp.~2383--2392, Association for Computational Linguistics, Nov. 2016.

\bibitem{stelmakh-etal-2022-asqa}
I.~Stelmakh, Y.~Luan, B.~Dhingra, and M.-W. Chang, ``{ASQA}: Factoid questions meet long-form answers,'' in {\em Proceedings of the 2022 Conference on Empirical Methods in Natural Language Processing} (Y.~Goldberg, Z.~Kozareva, and Y.~Zhang, eds.), (Abu Dhabi, United Arab Emirates), pp.~8273--8288, Association for Computational Linguistics, Dec. 2022.

\bibitem{wikians}
A.~Fader, L.~Zettlemoyer, and O.~Etzioni, ``{Open Question Answering Over Curated and Extracted Knowledge Bases},'' in {\em KDD}, 2014.

\bibitem{kocmi2022findings}
T.~Kocmi, R.~Bawden, O.~Bojar, A.~Dvorkovich, C.~Federmann, M.~Fishel, T.~Gowda, Y.~Graham, R.~Grundkiewicz, B.~Haddow, {\em et~al.}, ``Findings of the 2022 conference on machine translation (wmt22),'' in {\em Proceedings of the Seventh Conference on Machine Translation (WMT)}, pp.~1--45, 2022.

\bibitem{lin-2004-rouge}
C.-Y. Lin, ``{ROUGE}: A package for automatic evaluation of summaries,'' in {\em Text Summarization Branches Out}, (Barcelona, Spain), pp.~74--81, Association for Computational Linguistics, July 2004.

\bibitem{post-2018-call}
M.~Post, ``A call for clarity in reporting {BLEU} scores,'' in {\em Proceedings of the Third Conference on Machine Translation: Research Papers} (O.~Bojar, R.~Chatterjee, C.~Federmann, M.~Fishel, Y.~Graham, B.~Haddow, M.~Huck, A.~J. Yepes, P.~Koehn, C.~Monz, M.~Negri, A.~N{\'e}v{\'e}ol, M.~Neves, M.~Post, L.~Specia, M.~Turchi, and K.~Verspoor, eds.), (Brussels, Belgium), pp.~186--191, Association for Computational Linguistics, Oct. 2018.

\bibitem{panigrahi2019non}
A.~Panigrahi, R.~Somani, N.~Goyal, and P.~Netrapalli, ``Non-gaussianity of stochastic gradient noise,'' {\em arXiv preprint arXiv:1910.09626}, 2019.

\bibitem{anil2020scalable}
R.~Anil, V.~Gupta, T.~Koren, K.~Regan, and Y.~Singer, ``Scalable second order optimization for deep learning,'' {\em arXiv preprint arXiv:2002.09018}, 2020.

\bibitem{choudhary2020comprehensive}
T.~Choudhary, V.~Mishra, A.~Goswami, and J.~Sarangapani, ``A comprehensive survey on model compression and acceleration,'' {\em Artificial Intelligence Review}, vol.~53, pp.~5113--5155, 2020.

\bibitem{LeCun1989OptimalBD}
Y.~LeCun, J.~S. Denker, and S.~A. Solla, ``Optimal brain damage,'' in {\em Neural Information Processing Systems}, 1989.

\bibitem{Hanson1988ComparingBF}
S.~J. Hanson and L.~Y. Pratt, ``Comparing biases for minimal network construction with back-propagation,'' in {\em Neural Information Processing Systems}, 1988.

\bibitem{zhu2017prune}
M.~Zhu and S.~Gupta, ``To prune, or not to prune: exploring the efficacy of pruning for model compression,'' {\em arXiv preprint arXiv:1710.01878}, 2017.

\bibitem{oktay2019scalable}
D.~Oktay, J.~Ball{\'e}, S.~Singh, and A.~Shrivastava, ``Scalable model compression by entropy penalized reparameterization,'' {\em arXiv preprint arXiv:1906.06624}, 2019.

\bibitem{Haroush2019TheKW}
M.~Haroush, I.~Hubara, E.~Hoffer, and D.~Soudry, ``The knowledge within: Methods for data-free model compression,'' {\em 2020 IEEE/CVF Conference on Computer Vision and Pattern Recognition (CVPR)}, pp.~8491--8499, 2019.

\bibitem{polino2018model}
A.~Polino, R.~Pascanu, and D.~Alistarh, ``Model compression via distillation and quantization,'' {\em arXiv preprint arXiv:1802.05668}, 2018.

\bibitem{Han2015DeepCC}
S.~Han, H.~Mao, and W.~J. Dally, ``Deep compression: Compressing deep neural network with pruning, trained quantization and huffman coding,'' {\em arXiv: Computer Vision and Pattern Recognition}, 2015.

\bibitem{gray1984vector}
R.~Gray, ``Vector quantization,'' {\em IEEE Assp Magazine}, vol.~1, no.~2, pp.~4--29, 1984.

\bibitem{frantar2022gptq}
E.~Frantar, S.~Ashkboos, T.~Hoefler, and D.~Alistarh, ``Gptq: Accurate post-training quantization for generative pre-trained transformers,'' {\em arXiv preprint arXiv:2210.17323}, 2022.

\bibitem{lin2023awq}
J.~Lin, J.~Tang, H.~Tang, S.~Yang, X.~Dang, and S.~Han, ``Awq: Activation-aware weight quantization for llm compression and acceleration,'' {\em arXiv preprint arXiv:2306.00978}, 2023.

\bibitem{Sharma2023TheTI}
P.~Sharma, J.~T. Ash, and D.~Misra, ``The truth is in there: Improving reasoning in language models with layer-selective rank reduction,'' {\em ArXiv}, vol.~abs/2312.13558, 2023.

\bibitem{yadav2023compeft}
P.~Yadav, L.~Choshen, C.~Raffel, and M.~Bansal, ``Compeft: Compression for communicating parameter efficient updates via sparsification and quantization,'' {\em arXiv preprint arXiv:2311.13171}, 2023.

\bibitem{aghajanyan-etal-2021-intrinsic}
A.~Aghajanyan, S.~Gupta, and L.~Zettlemoyer, ``Intrinsic dimensionality explains the effectiveness of language model fine-tuning,'' in {\em Proceedings of the 59th Annual Meeting of the Association for Computational Linguistics and the 11th International Joint Conference on Natural Language Processing (Volume 1: Long Papers)} (C.~Zong, F.~Xia, W.~Li, and R.~Navigli, eds.), (Online), pp.~7319--7328, Association for Computational Linguistics, Aug. 2021.

\bibitem{gueta-etal-2023-knowledge}
A.~Gueta, E.~Venezian, C.~Raffel, N.~Slonim, Y.~Katz, and L.~Choshen, ``Knowledge is a region in weight space for fine-tuned language models,'' in {\em Findings of the Association for Computational Linguistics: EMNLP 2023} (H.~Bouamor, J.~Pino, and K.~Bali, eds.), (Singapore), pp.~1350--1370, Association for Computational Linguistics, Dec. 2023.

\bibitem{ilharco2022editing}
G.~Ilharco, M.~T. Ribeiro, M.~Wortsman, S.~Gururangan, L.~Schmidt, H.~Hajishirzi, and A.~Farhadi, ``Editing models with task arithmetic,'' {\em arXiv preprint arXiv:2212.04089}, 2022.

\bibitem{Choshen2022FusingFM}
L.~Choshen, E.~Venezian, N.~Slonim, and Y.~Katz, ``Fusing finetuned models for better pretraining,'' {\em ArXiv}, vol.~abs/2204.03044, 2022.

\bibitem{Wortsman2022ModelSA}
M.~Wortsman, G.~Ilharco, S.~Y. Gadre, R.~Roelofs, R.~Gontijo-Lopes, A.~S. Morcos, H.~Namkoong, A.~Farhadi, Y.~Carmon, S.~Kornblith, and L.~Schmidt, ``Model soups: averaging weights of multiple fine-tuned models improves accuracy without increasing inference time,'' 2022.

\bibitem{matena2021merging}
M.~Matena and C.~Raffel, ``Merging models with fisher-weighted averaging,'' {\em arXiv preprint arXiv:2111.09832}, 2021.

\bibitem{yadav2023ties}
P.~Yadav, D.~Tam, L.~Choshen, C.~Raffel, and M.~Bansal, ``Ties-merging: Resolving interference when merging models,'' in {\em Thirty-seventh Conference on Neural Information Processing Systems}, 2023.

\bibitem{zhang2023adaptive}
Q.~Zhang, M.~Chen, A.~Bukharin, P.~He, Y.~Cheng, W.~Chen, and T.~Zhao, ``Adaptive budget allocation for parameter-efficient fine-tuning,'' {\em arXiv preprint arXiv:2303.10512}, 2023.

\bibitem{dettmers2023qlora}
T.~Dettmers, A.~Pagnoni, A.~Holtzman, and L.~Zettlemoyer, ``Qlora: Efficient finetuning of quantized llms,'' {\em arXiv preprint arXiv:2305.14314}, 2023.

\bibitem{sabne2020xla}
A.~Sabne, ``Xla: Compiling machine learning for peak performance,'' 2020.

\bibitem{wu2023pytorch}
P.~Wu, ``Pytorch 2.0: The journey to bringing compiler technologies to the core of pytorch (keynote),'' in {\em Proceedings of the 21st ACM/IEEE International Symposium on Code Generation and Optimization}, pp.~1--1, 2023.

\bibitem{tyagi2023gravac}
S.~Tyagi and M.~Swany, ``Gravac: Adaptive compression for communication-efficient distributed dl training,'' in {\em 2023 IEEE 16th International Conference on Cloud Computing (CLOUD)}, pp.~319--329, IEEE, 2023.

\bibitem{zhao2024galore}
J.~Zhao, Z.~Zhang, B.~Chen, Z.~Wang, A.~Anandkumar, and Y.~Tian, ``Galore: Memory-efficient llm training by gradient low-rank projection,'' {\em arXiv preprint arXiv:2403.03507}, 2024.

\bibitem{radford2019better}
A.~Radford, J.~Wu, D.~Amodei, D.~Amodei, J.~Clark, M.~Brundage, and I.~Sutskever, ``Better language models and their implications,'' {\em OpenAI blog}, vol.~1, no.~2, 2019.

\bibitem{wav2vec19}
S.~Schneider, A.~Baevski, R.~Collobert, and M.~Auli, ``{wav2vec: Unsupervised Pre-Training for Speech Recognition},'' in {\em Proc. Interspeech 2019}, pp.~3465--3469, 2019.

\bibitem{liu2019RoBERTa}
Y.~Liu, M.~Ott, N.~Goyal, J.~Du, M.~Joshi, D.~Chen, O.~Levy, M.~Lewis, L.~Zettlemoyer, and V.~Stoyanov, ``Roberta: A robustly optimized bert pretraining approach,'' {\em arXiv preprint arXiv:1907.11692}, 2019.

\bibitem{radford2021learning}
A.~Radford, J.~W. Kim, C.~Hallacy, A.~Ramesh, G.~Goh, S.~Agarwal, G.~Sastry, A.~Askell, P.~Mishkin, J.~Clark, {\em et~al.}, ``Learning transferable visual models from natural language supervision,'' in {\em International conference on machine learning}, pp.~8748--8763, PMLR, 2021.

\bibitem{almazrouei2023falcon}
E.~Almazrouei, H.~Alobeidli, A.~Alshamsi, A.~Cappelli, R.~Cojocaru, M.~Debbah, {\'E}.~Goffinet, D.~Hesslow, J.~Launay, Q.~Malartic, {\em et~al.}, ``The falcon series of open language models,'' {\em arXiv preprint arXiv:2311.16867}, 2023.

\bibitem{workshop2022bloom}
B.~Workshop, T.~L. Scao, A.~Fan, C.~Akiki, E.~Pavlick, S.~Ili{\'c}, D.~Hesslow, R.~Castagn{\'e}, A.~S. Luccioni, F.~Yvon, {\em et~al.}, ``Bloom: A 176b-parameter open-access multilingual language model,'' {\em arXiv preprint arXiv:2211.05100}, 2022.

\bibitem{openlm2023openllama}
X.~Geng and H.~Liu, ``Openllama: An open reproduction of llama,'' May 2023.

\bibitem{touvron2023llama}
H.~Touvron, L.~Martin, K.~Stone, P.~Albert, A.~Almahairi, Y.~Babaei, N.~Bashlykov, S.~Batra, P.~Bhargava, S.~Bhosale, {\em et~al.}, ``Llama 2: Open foundation and fine-tuned chat models,'' {\em arXiv preprint arXiv:2307.09288}, 2023.

\bibitem{ivison2023camels}
H.~Ivison, Y.~Wang, V.~Pyatkin, N.~Lambert, M.~Peters, P.~Dasigi, J.~Jang, D.~Wadden, N.~A. Smith, I.~Beltagy, and H.~Hajishirzi, ``Camels in a changing climate: Enhancing lm adaptation with tulu 2,'' 2023.

\end{thebibliography}
\bibliographystyle{ieeetr}

\appendix





\subsection{Models name Downloaded from Hugging Face}\label{app:all_models}
The main models used are:
For FP32 Models: Bert~\cite{Devlin2019BERTPO}, GPT2~\cite{radford2019better}, wav2Vec~\cite{wav2vec19}, RoBERTa~\cite{liu2019RoBERTa} and fine tuned RoBERTa on the Rotten Tomato dataset~\cite{rottenTomatos}, XLM-Roberta,  CLIP~\cite{radford2021learning}, and T5-base~\cite{raffel2020exploring}. 

For BF16 Models: Falcon-7B~\cite{almazrouei2023falcon}, Bloom~\cite{workshop2022bloom} and OpenBuddy/OPENLLAMA (3B-BF16,~\cite{openlm2023openllama}) and Mistral~\cite{jiang2023mistral}.
For FP16 Models: llama2-13B \cite{touvron2023llama} Tulu-7B \cite{ivison2023camels} and argilla/CapybaraHermes-2.5-Mistral-7B~\cite{jiang2023mistral} and Stable-Video-Diffusion. 

We provide a full list of the exact model names used as they appear in Hugging~Face  hub for complete reproducibility. 

\begin{itemize}
   \item jonatasgrosman/wav2vec2-large-xlsr-53-english
   \item google-bert/bert-base-uncased
   \item openai-community/gpt2
    \item runwayml/stable-diffusion-v1-5
   \item becausecurious/stable-video-diffusion-img2vid-fp16
   \item argilla/CapybaraHermes-2.5-Mistral-7B
   \item FacebookAI/roberta-base
   \item FacebookAI/xlm-roberta-base
   \item openai/clip-vit-large-patch14
   \item google-t5/t5-base
   \item TheBloke/Llama-2-13B-Chat-fp16
   \item TheBloke/tulu-7B-fp16 
   \item tiiuae/falcon-7b
   \item bigscience/bloom 
   \item OpenBuddy/openbuddy-openllama-3b-v10-bf16
   \item mistralai/Mistral-7B-v0.1
   
\end{itemize}

\subsection {Quantized Models} \label{app:quantized}

We provide the names of the Quantized Models used as they appear in Hugging~Face hub for complete reproducibility. We discuss compressing those in section~\ref{sec:related}.

\begin{itemize}
    \item TheBloke/CapybaraHermes-2.5-Mistral-7B-GPTQ
    \item TheBloke/CapybaraHermes-2.5-Mistral-7B-GPTQ
    \item TheBloke/CapybaraHermes-2.5-Mistral-7B-GPTQ
    \item TheBloke/CapybaraHermes-2.5-Mistral-7B-GPTQ
    \item TheBloke/CapybaraHermes-2.5-Mistral-7B-GPTQ
    \item TheBloke/CapybaraHermes-2.5-Mistral-7B-AWQ
\end{itemize}

\subsection{RoBERTa-base model trained on tweets} \label{app:deltas_models}
We provide the names of the models finetuned on twitter dataset, as they appear on Hugging~Face hub for complete reproducibility. We discuss compressing those in \S\ref{sec:deltas}.
\begin{itemize}
     \item cardiffnlp/twitter-roberta-base-irony 
     \item cardiffnlp/twitter-roberta-base-offensive 
     \item cardiffnlp/twitter-roberta-base-hate 
     \item cardiffnlp/twitter-roberta-base-mar2020
     \item cardiffnlp/twitter-roberta-base-jun2020   
     \item cardiffnlp/twitter-roberta-base-sep2020    
     \item cardiffnlp/twitter-roberta-base-dec2020     
     \item cardiffnlp/twitter-roberta-base-mar2021
     \item cardiffnlp/twitter-roberta-base-jun2021   
     \item cardiffnlp/twitter-roberta-base-sep2021    
     \item cardiffnlp/twitter-roberta-base-dec2021     
\end{itemize}

\end{document}